\documentclass[sigconf]{acmart}

\renewcommand\footnotetextcopyrightpermission[1]{} % removes footnote with conference information in first column
\AtBeginDocument{%
  }

\usepackage{amsmath}
\usepackage{mathtools}
\usepackage{algorithm}

\usepackage{xspace}
\usepackage{color}
\usepackage{algpseudocode}
\usepackage{graphicx}
\usepackage{subcaption} % For subfigures
\usepackage{float}      % For more control over figure positioning

\usepackage{etoolbox}
\usepackage{tikz}
\usetikzlibrary{tikzmark}
\usetikzlibrary{calc}

\errorcontextlines\maxdimen

\newcommand{\ALGtikzmarkcolor}{black}% Customize this, if you want
\newcommand{\ALGtikzmarkextraindent}{4pt}% Customize this, if you want
\newcommand{\ALGtikzmarkverticaloffsetstart}{-.5ex}% Customize this, if you want
\newcommand{\ALGtikzmarkverticaloffsetend}{-.5ex}% Customize this, if you want

\makeatletter
\newcounter{ALG@tikzmark@tempcnta}

\newcommand\ALG@tikzmark@start{%
    \global\let\ALG@tikzmark@last\ALG@tikzmark@starttext%
    \expandafter\edef\csname ALG@tikzmark@\theALG@nested\endcsname{\theALG@tikzmark@tempcnta}%
    \tikzmark{ALG@tikzmark@start@\csname ALG@tikzmark@\theALG@nested\endcsname}%
    \addtocounter{ALG@tikzmark@tempcnta}{1}%
}

\def\ALG@tikzmark@starttext{start}
\newcommand\ALG@tikzmark@end{%
    \ifx\ALG@tikzmark@last\ALG@tikzmark@starttext
    \else
        \tikzmark{ALG@tikzmark@end@\csname ALG@tikzmark@\theALG@nested\endcsname}%
        \tikz[overlay,remember picture] \draw[\ALGtikzmarkcolor] let \p{S}=($(pic cs:ALG@tikzmark@start@\csname ALG@tikzmark@\theALG@nested\endcsname)+(\ALGtikzmarkextraindent,\ALGtikzmarkverticaloffsetstart)$), \p{E}=($(pic cs:ALG@tikzmark@end@\csname ALG@tikzmark@\theALG@nested\endcsname)+(\ALGtikzmarkextraindent,\ALGtikzmarkverticaloffsetend)$) in (\x{S},\y{S})--(\x{S},\y{E});%
    \fi
    \gdef\ALG@tikzmark@last{end}%
}

\apptocmd{\ALG@beginblock}{\ALG@tikzmark@start}{}{\errmessage{failed to patch}}
\pretocmd{\ALG@endblock}{\ALG@tikzmark@end}{}{\errmessage{failed to patch}}
\makeatother
\algdef{SE}[SUBALG]{Indent}{EndIndent}{}{\algorithmicend\ }%
\algtext*{Indent}
\algtext*{EndIndent}
\setcopyright{acmlicensed}
\copyrightyear{2018}
\acmYear{2018}
\acmDOI{XXXXXXX.XXXXXXX}
\acmConference[Conference acronym 'XX]{Make sure to enter the correct
  conference title from your rights confirmation emai}{June 03--05,
  2018}{Woodstock, NY}
\acmISBN{978-1-4503-XXXX-X/18/06}

\DeclarePairedDelimiter\norm{\lVert}{\rVert}

\makeatletter
\renewcommand\@formatdoi[1]{\ignorespaces}
\makeatother

\begin{document}

\title{Copyright-Aware Incentive Scheme for Generative Art Models Using Hierarchical Reinforcement Learning}

\author{Zhuan Shi}
\affiliation{%
  \institution{EPFL}
  \city{Lausanne}
  \country{Switzerland}
}
\email{zhuan.shi@epfl.ch}

\author{Yifei Song}
\affiliation{%
  \institution{The Chinese University of Hong Kong}
  \city{Hong Kong}
  \country{China}
}
\email{christine.yifei.song@outlook.com}

\author{Xiaoli Tang}
\affiliation{%
  \institution{Nanyang Technological University}
  \city{}
  \country{Singapore}
}
\email{xiaoli001@ntu.edu.sg}

\author{Lingjuan Lyu}
\affiliation{%
  \institution{Sony AI}
  \city{Zurich}
  \country{Switzerland}
}
\email{Lingjuan.Lv@sony.com}

\author{Boi Faltings}
\affiliation{%
  \institution{EPFL}
  \city{Lausanne}
  \country{Switzerland}
}
\email{boi.faltings@epfl.ch}

\begin{abstract}
Generative art using Diffusion models has achieved remarkable performance in image generation and text-to-image tasks. However, the increasing demand for training data in generative art raises significant concerns about copyright infringement, as models can produce images highly similar to copyrighted works. Existing solutions attempt to mitigate this by perturbing Diffusion models to reduce the likelihood of generating such images, but this often compromises model performance. Another approach focuses on economically compensating data holders for their contributions, yet it fails to address copyright loss adequately. Our approach begin with the introduction of a novel copyright metric grounded in copyright law and court precedents on infringement. We then employ the TRAK method to estimate the contribution of data holders. To accommodate the continuous data collection process, we divide the training into multiple rounds. Finally, We designed a hierarchical budget allocation method based on reinforcement learning to determine the budget for each round and  the remuneration of the data holder based on the data holder's contribution and copyright loss in each round. Extensive experiments across three datasets show that our method outperforms all eight benchmarks, demonstrating its effectiveness in optimizing budget distribution in a copyright-aware manner. To the best of our knowledge, this is the first technical work that introduces to incentive contributors and protect their copyrights by compensating them.
\end{abstract}

\keywords{Copyright Protection, Incentive Scheme, Diffusion Model}

\makeatletter
\def\@copyrightspace{\relax}
\makeatother

 \setcopyright{none}

\maketitle
% \textbf{ACM Reference Format:}\\
% The Name of the Title Is Hope. In \textit{Proceedings of Make sure to enter the correct conference title from your rights confirmation email (Conference acronym 'XX')}, June 03--05, 2018, Woodstock, NY, USA. ACM, New York, NY, USA, 6 pages. https://doi.org/XXXXXXX.XXXXX

\pagestyle{plain}

\fancyfoot{}

\thispagestyle{empty}

\section{Introduction} 
% describe the popularity of generative models and then point out the diffusion model
% The recent years have witnessed a considerable growth in the generative models, and diffusion models such as Dall$\cdot$E \cite{dalle} and Stable Diffusion \cite{sdxl} have demonstrated impressive performance in generation tasks and other downstream tasks.

% Recently, generative art as a typical category of artificial intelligence-generated content (AIGC), has become a cutting-edge research topic \cite{aigc}\cite{ai_art_impact}. Copyright-aware incentive scheme for generative models are crucial for the web as they ensure compliance with intellectual property laws, promoting ethical content generation and fostering a better ecosystem for generative arts on the Internet. With the development of diffusion models (such as  Dall$\cdot$E \cite{dalle} and Stable Diffusion \cite{sdxl}), generative art has achieved extremely impressive performance in image generation or text-to-image tasks.

Generative models are playing an increasingly important role in the web ecosystem by revolutionizing the way digital content is created and consumed \cite{wu2024smalllanguagemodelsserve} \cite{Mechanism_design_LLM}. 
These models enable the automatic generation of high-quality images, audio, and video, leading to more dynamic, personalized, and engaging web experiences \cite{vombatkere2024tiktokart}. 
With the rise of generative models, generative art, a prominent form of artificial intelligence-generated content (AIGC), has emerged as a cutting-edge research topic \cite{aigc}\cite{ai_art_impact}. With the advancements in diffusion models, such as DALL·E \cite{dalle} and Stable Diffusion \cite{sdxl}, generative art has demonstrated remarkable progress, particularly in image generation and text-to-image tasks.

% introduce the copyright issues in diffusion models and describe the existing mehtods for addressing the copyright issues. describe the drawback of the first methods and point out the strength of the second methods(using economic solution).

The widespread deployment of such generative art models has brought about significant challenges concerning the risk for copyright infringement. The image generative models require large amount of training data, and it's impractical to only use non-copyrighted content to train the models. The model holder wants to leverage the high quality copyrighted data to learn the style and content and output high quality generated data. However, the generative models may generate some images that are very similar to the copyrighted training data, which may lead to copyright infringement \cite{copyright_in_generative_ai}.

% Existing methods to solve copyright issues mainly come from two aspects:
% (1) perform perturbations to the training process of the model and lower the probability of generating similar images as the copyrighted training data; (2) preserve the original generative model and pay the copyrighted data holder a sensible amount of fee. Most previous works \cite{vyas2023provable} \cite{maini2023neural_memorization} \cite{xu2024pac} focused on the first approach, but we found the second one more desirable for the following reasons: (1) the perturbations added on the generative models often have negative impact on the quality of the model output, which will make the model less competitive; (2) the fee paying economic solution can encourage the data holder to provide more original and high-quality content, and thus foster a positive ecosystem for the visual art market.

Various methods have been developed to protect copyright in source data, with the most common strategy involving the introduction of perturbations during the training process to effectively safeguard the dataset’s copyright:
% for protecting copyright in source data have been developed, with the primary approach being to introduce perturbations during the training process to safeguard copyright dataset effectively: 
(1) Unrecognizable examples \cite{gandikota2023erasing, zhang2023forgetmenot} hinder models from learning essential features of protected images, either at the inference or training stages. However, this method is highly dependent on the specific images and models involved, and it lacks universal guarantees.
(2) Watermarking \cite{dogoulis2023improving, epstein2023online} embeds subtle, imperceptible patterns into images to detect copyright violations, though further research is needed to improve its reliability. 
(3) Machine unlearning \cite{bourtoule2020machine, huang2021unlearnable, gao2023implicit, nguyen2022survey} removes the influence of copyrighted data, supporting the right to be forgotten.
(4) Dataset deduplication \cite{somepalli2022diffusion} helps reduce memorization of training samples, minimizing the risk of copying protected content.

% Despite these efforts, existing copyright protection methods still have the following limitations: 
% (1) they lack a copyright metric that complies with copyright laws and regulations, which can be used to determine whether the generated images infringe copyright; (2) these methods that introduce perturbations into generative models to protect data copyright often negatively impact the quality of the model's output, rendering the model less competitive. 

Despite these efforts, existing copyright protection techniques that introduce perturbations into generative art models to safeguard data often come at the cost of output quality, reducing the model's competitiveness. In real-world applications, generative model developers prioritize the quality of the generated content, and any compromise in this area could result in significant financial losses, potentially amounting to billions of dollars \cite{gong2020survey}. 

\begin{figure}[htbp]
    \centering
    \includegraphics[width=\columnwidth]{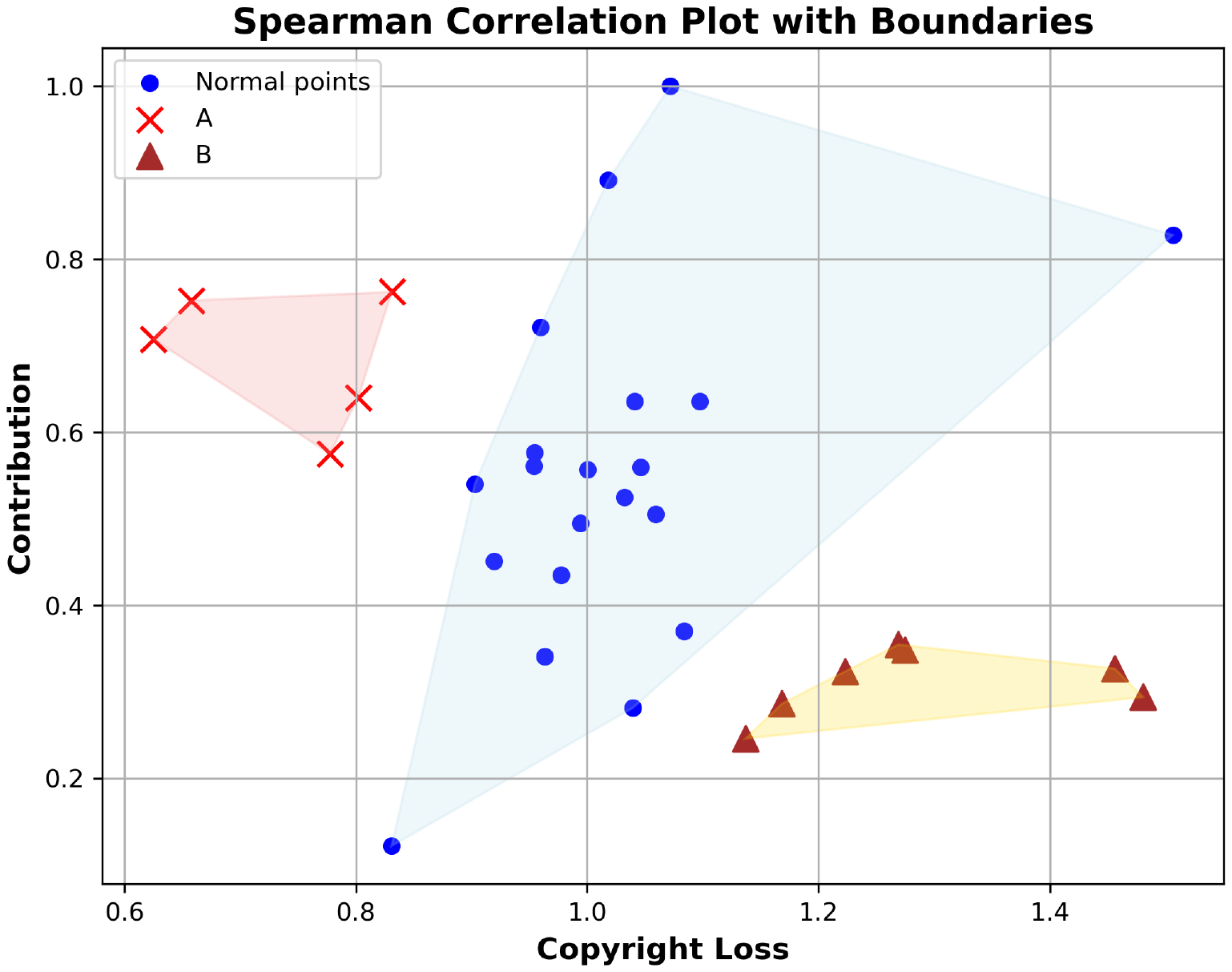}
    \vspace{-0.3in}
    \caption{The Results of the Spearman's Rank Correlation Test for Copyright Loss and Contribution indicate distinct group behaviors: Points in group A exhibit small copyright loss and significant contributions, while points in group B show minimal contribution but substantial copyright loss.}
    \label{fig:correlation}
\end{figure}

Therefore, some researchers have begun investigating economic solutions that provide fair compensation to copyright holders while maintaining the data quality of generative models.
For instance, \cite{economicSolutionCopyright} proposes treating the copyright loss of a data point as equivalent to its contribution. This approach introduces a framework where copyright owners are compensated proportionally to the contribution of their data, utilizing a royalty distribution model grounded in the probabilistic nature of AI and principles of cooperative game theory.
% This approach allows for fair compensation and incentivizes the provision of high-quality training data while maintaining the full capabilities of AI models.
However, In some cases, copyright loss and contribution are not the same concept. For example, some open source training data can have high contribution to the model, while the copyright loss is relatively low. On the contrary, some training data like trademarks can have low contribution to the model, while the copyright loss is relatively high when the generated content mimics some elements or idea of them. We conducted an experiment to illustrate this situation, where we use TRAK score as the contribution (see definition in Sec. \ref{subsec:contribution}) and use the proposed metric as copyright loss (see definition in Sec. \ref{subsec:copyright}). As shown in Figure \ref{fig:correlation}, points in group A (red crosses) are those that have small copyright loss and big contribution to the model, while points in group B (brown triangles) are those that have big copyright loss and small contribution to the model.

Based on the above experimental observations, we hope to design an incentive scheme for the model holder, which can compensate data holders based on their contributions and copyright losses during the training process. In real-world applications, the model holder may have a total budget for training the generative model. In addition, during training, new data will be collected continuously. Due to the continuous data collection process, there would be multiple iterative versions of the generative model \cite{wang2024comprehensivesurveycontinuallearning}. 

In this context, we encounter several challenges when designing incentive schemes: (1) The absence of a standardized copyright metric that complies with existing copyright laws and regulations complicates the determination of whether generated images infringe on copyrights; (2) Allocating the total budget across various versions of the generated model poses difficulties. Investing excessively in earlier versions may leave new data holders, who join at a later stage, without adequate compensation.

To tackle these challenges, we introduce a Copyright-aware Incentive Scheme tailored for generative art models. Drawing inspiration from U.S. court practices \cite{Copyrightlaw1989}, which utilize a two-part test to assess copyright violations, we develop a copyright metric that aligns with these legal standards. This metric encompasses an extrinsic component that evaluates objective similarity in specific expressive elements, as well as an intrinsic component that considers subjective similarity from the perspective of a reasonable audience. By integrating both semantic and perceptual similarity, our proposed metric effectively reflects judicial criteria. Additionally, we have designed a hierarchical budget allocation method based on reinforcement learning, which first establishes the budget for each round and subsequently determines the remuneration for data holders based on their contributions and the associated copyright loss in each round.

% our contributions are as follows:
Our contribution are as follows:
\begin{enumerate}
    \item We propose the first metric for copyright loss that closely follows the procedures of copyright law.
    \item We designed a hierarchical budget allocation method based on reinforcement learning, which first determines the budget for each round and then determines the remuneration of the data holder based on the data holder's contribution and copyright loss in each round. This is the first incentive scheme that compensates data contributors based on copyright loss and contributions.
    \item Extensive experimental comparisons between existing incentive schemes and our method demonstrate that our approach is more attractive to data holders who possess high-quality and original images.
\end{enumerate}

\section{Related works}
\paragraph{Text-to-image Models} Generative Adversarial Networks (GAN) \cite{reed2016generative} are one of the early approaches to generating images from text. Following works like StackGAN \cite{stackgan2017}, AttnGAN \cite{xu2018attngan} improve the quality and relevance of generated images of the text-to-image models. These year, diffusion models have become a prominent choice in generative modeling \cite{generative_diffusion_model_survey}. Diffusion models are a kind of generative models that differ from the traditional models like GANs in that they iteratively transform noisy images to recover the original images through a sequence of learned denoising steps \cite{ho2020denoising}. \cite{ldm} introduces Latent Diffusion Models (LDMs), which combines diffusion processes with latent variable modeling. The authors proposed a model that operates in a lower-dimensional latent space rather than pixel space. This approach significantly reduces computational costs and enables the generation of high-resolution images with less resource usage. The diffusion models have shown remarkable abilities in generating diverse and high-quality outputs.

\paragraph{Copyright Protection} Copyright is crucial in AI-generated images to protect intellectual property, ensure fair use, and encourage innovation by legally securing the rights of creators and developers. In both US and EU law \cite{Copyrightlaw1989}\cite{EUR}, the 
substantiality of copyright infringement is one of the most important and measurable determinants. In practice, the court will perform extrinsic and intrinsic tests to measure the substantiality \cite{Sid&Marty_1977} and determine whether the infringement exists. There are some previous works focusing on building a generative model that avoids generating content mimic the copyrighted images. Existed method are machine unlearning, and fine-tuning. For example, "Forgot-Me-Not" \cite{zhang2023forgotmenot} is a method that can safely remove specified IDs, objects, or styles from a well-configured text-to-image model in as little as 30 seconds. \cite{vyas2023provable} proposes the concept "Near Access Free(NAF)" and also a method that can output generative models with strong bounds on the probability of sampling protected content.
% Preliminary experiment here. Explain the phenomenon.
% \paragraph{Data Attribution} Data Attribution refers to the task of measuring the influence of a data point in the model training set to the model output \cite{koh2017understanding}. Some widely used data attribution algorithms are Leave-One-Out, Shapley Value, Influence function. Many of the widely used data attribution algorithms suffer from high computation cost, and some derivative works have proposed algorithm that improve time efficiency, e.g. \cite{profitAllocShapley}, \cite{wei2020efficient}, \cite{DynamicShapley}. 
\paragraph{Reinforcement Learning} Reinforcement Learning is a popular pivotal area in machine learning, addressing the challenge of learning optimal actions through interaction with an environment. There are three basic types of RL: value-based, policy-based, and actor-critic. Q-learning is a value-based and model-free algorithm that seeks to learn the value of actions in various states, eventually deriving an optimal policy. By integrating deep learning with Q-learning, Deep Q-Network (DQN) \cite{DQN2015} enabled RL algorithms to handle high-dimensional state spaces,  which demonstrated RL's potential to solve complex problems previously considered intractable. In contrast to value-based methods, policy-based approaches focus directly on learning the policy itself. Recent advancements include the development of Proximal Policy Optimization (PPO) \cite{schulman2017ppo}, which addresses some limitations of earlier policy gradient methods. PPO improves stability and performance by using clipped objective functions and a robust optimization approach, making it highly effective for various RL tasks.

\section{Preliminaries and Problem Formulation}
In this section, we will introduce the preliminaries and the problem formulation. 
The notations used in this paper are listed in Table.\ref{notation} for ease of reference. 

\subsection{Preliminaries}

% Notation Form
\begin{table}[t]
\centering
\small
\caption{Notations.}
\resizebox{0.47\textwidth}{!}{
\begin{tabular}{|c|l|}
\hline
\textbf{Symbol} & \textbf{Description} \\ \hline
$\mathcal{M}_t$ & The model at round $t$ \\ \hline
$B, B^t$ & Total budget, the budget allocated to round $t$ \\ \hline
% $B^t$ & The budget allocated to round $t$\\ \hline
$\mathcal{D}_k$ & The dataset of data holder $k$ \\ \hline
$\theta$ & The parameter set of the model \\ \hline
$c_i$ & Comprehensive similarity of sample i \\ \hline
$\hat{c^t_i}$ & Comprehensive copyright loss of training sample $i$ at round $t$\\ \hline
$c^t_k$ & Copyright loss of data holder $k$ at round $t$\\ \hline
$\beta^t_i$ & Contribution of training sample $i$ at round $t$\\ \hline
$\beta^t_k$ & Contribution of data holder $k$ at round $t$\\ \hline
$\hat{\beta}^t_i$ & Normalized contribution of training sample $i$ at round $t$\\ \hline
$a,b$ & Coefficients of copyright loss\\ \hline
$Dist_{sem}$ & Semantic distance \\ \hline
$Dist_{per}$ & Perceptual distance \\ \hline
$\mathcal{A}_t$ & The action of the outer RL at round $t$\\ \hline
$\mathcal{R}_t$ & The reward of the outer RL at round $t$\\ \hline
$r_t$ & The reward of the inner RL at round $t$ \\ \hline
$\lambda, \delta$ & Coefficients of the reward of the inner RL\\ \hline
$p^t$ & The budget allocation vector at round $t$\\ \hline
$P^t_k$ & Payoff of the data holder $k$ at round $t$\\ \hline
\end{tabular}}
\label{notation}
\end{table}
%%%%%%%%%%%%%

\begin{definition}[Text-to-image diffusion models]
Text-to-image diffusion models represent a category of algorithms that systematically introduce noise into input images to create noised samples, then reverse the process by learning to remove noise with a guided prompt embedding at each step, ultimately working towards reconstructing an image according to the prompt \cite{diffusion_model_survey}.

The forward process that transforms the input $x_0$ into a prior distribution, which can be written as \cite{sdxl}:
\begin{equation}
q(\mathbf{x}_1, \dots, \mathbf{x}_\mathcal{T} \mid \mathbf{x}_0) = \prod_{t=1}^{\mathcal{T}} q(\mathbf{x}_v \mid \mathbf{x}_{v-1}),
\end{equation}
where $x_v$ is the sample at time step $v\in[1,\mathcal{}{\mathcal{T}}]$.
The kernel $q(\mathbf{x}_v \mid \mathbf{x}_{v-1})$ is often defined as:
\begin{equation}
q(\mathbf{x}_v \mid \mathbf{x}_{v-1}) = \mathcal{N}(\mathbf{x}_v; \sqrt{1 - \gamma_v} \mathbf{x}_{v-1}, \gamma_t \mathbf{I}),
\end{equation}
where $\gamma_v \in (0,1)$ is a hyperparameter that controls the variance of the noise added at each time step in the diffusion process.
The model can denoise the noise image $x_v$ and get back to $x_0$ by iteratively sampling from the transition kernel, where the transition kernel is defined as:

\begin{equation}
p_{\theta}(\mathbf{x}_{v-1} \mid \mathbf{x}_v) = \mathcal{N}(\mathbf{x}_{v-1}; \mu_{\theta}(\mathbf{x}_v, v), \Sigma_{\theta}(\mathbf{x}_v, v)),
\end{equation}
where $\mu_{\theta}(\mathbf{x}_v, v)$ and $\Sigma_{\theta}(\mathbf{x}_v, v)$ are the mean and the covariance of the Gaussian distribution, correspondingly.

We train the transition kernel so that $p_{\theta}(x_v)$ can closely approximate the true data distribution $q(x_0)$. Then the training objective that we want to minimize is defined as follows \cite{diffusion_model_survey}:
% \begin{equation}
% \mathbb{E}_{\mathbf{x}_0 \sim q(\mathbf{x}_0)},t
% \left[ \|\boldsymbol{\epsilon} - \boldsymbol{\epsilon}_{\theta}(\mathbf{x}_t, t)\|^2 \right]
% \end{equation}
\begin{equation}
    \mathcal{L}_{TIDM}:=\mathbb{E}_{\epsilon,v}\norm{\epsilon-\epsilon_{\theta}(z_v,v, \mathcal{E}(l))}^{2}_2,    
\end{equation}
where $z_v$ is the latent representation of the $x_v$ in the forward process, $\mathcal{E}$ is a text encoder for the guiding prompt input $l$.
% where and $\epsilon$ follows the Gaussian distribution, and 
% \(\overline{\alpha_v}\) is defined as:

% \begin{equation}
% \overline{\alpha_v} = \prod_{s=1}^v (1 - \beta_s),
% \end{equation}

\end{definition}

\begin{definition}[Data attribution method: TRAK]
\label{def: trak}
TRAK \cite{park2023trak} is an efficient data attribution method to attribute both linear and non-linear models. Given a training set $D$, randomly sample $N$ subsets $\{D^1, D^s, ..., D^N\}$ with a fix size $K$ from the training set $D$. The optimal model parameter set on training set $D^s$ is denoted as $\theta^*_s$. A random projection matrix $P$ is used in reducing the dimension.

The data attribution score for each sample $i$ using TRAK algorithm is formulated as follows \cite{zheng2024dtrak}:
\begin{equation}
    \label{eq: trak}
    \tau_{TRAK}(x) = ( \frac{1}{N} \sum_{s=1}^{N} \phi^s(x)^\top \left( \Phi^s{}^\top \Phi^s \right)^{-1} \Phi^s{}^\top )( \frac{1}{N} \sum_{s=1}^{N} \mathcal{Q}^s).
\end{equation}

$\Phi^s$ is the projected gradient matrix:
\begin{equation}
    \Phi^s = \left[ \phi^s(x^1); \cdots ; \phi^s(x^K) \right]^\top,
\end{equation}
where $\phi^s(x) = \mathbf{P}_s^\top \nabla_{\theta} \mathcal{F}(x; \theta_s^*)$ is the projected gradient.

$\mathcal{Q}^s$ is the weighting term:
\begin{equation}
    \mathcal{Q}^{s} = \text{diag} \left( Q^s(x^1), \cdots, Q^s(x^N) \right).
\end{equation}
where $Q^s(x) = \frac{\partial \mathcal{L}}{\partial \mathcal{F}}(x; \theta_s^*)$
, $\mathcal{L}$ is the loss function or the training objective of the model, and $\mathcal{F}$ is the output function of the model.
\end{definition}

\subsection{Problem Formulation}
%版本问题 continuting learning
Considering the generative model training process, a model holder needs data to train the generative model, a group of data holders with data $\mathcal{D}_k$ join the training process, and they perform the gradient updates and finetune the model. 

% \note{describe why exist many rounds, just like introduced in intruduction}
The training time of the generative model could be long, and at each period the joining data holders can be different.
The model will be updated at each round $t\in(1,T)$, and we denoted the output model after the training process of round $t$ as model version $t$  $\mathcal{M}_t$. Denote the training set of model $M_t$ by $\mathcal{D}_t = ({\mathcal{D}}_1, ..., \mathcal{D}_k, ..., \mathcal{D}_n)$, and the training sample $i$ by $x_i\in\mathcal{D}_k$. $\theta_{S}$ is the parameter set of the model with the training set $S$, where $S \subseteq \{\mathcal{D}_1, ..., \mathcal{D}_k, ..., \mathcal{D}_n\}$. The model holder has a total budget $B$, we would like to distribute the budget to the data holder $k$ in an incentive way so that the output model performance could be optimized. The problems of budget distribution naturally arise during the training process with multiple data holders: (1) how to distribute budget among the different rounds of training; (2) how to distribute budget among the data holders at each round, so that the data holders are paid fairly based on the "contribution" and "copyright loss" they have in the training? Our goal is to build an incentive scheme to assign a sensible payment $P^t_k$ to each data holder $k$ at round $t$ by quantifying the contribution and the copyright loss of the training sample in the training process.

To achieve our goal, we divide the whole problem into the following problems:
\begin{enumerate}
    \item \textbf{Contribution Assessment:} 
    The contribution of the data points is crucial in distributing the total budget $B$. The complexity of the generative models, especially diffusion models in our case, puts barriers on computing the contribution of each training data point. Traditional approaches of contribution assessment require multiple times of retraining, which is very unfeasible for big generative models like Stable Diffusion \cite{diffusion_model}.
    % to give a quantified estimation of the "impact" of a particular training sample in the training set, we leverage the concepts and tools in $\it{Data}$ $\it{Attribution}$. We use the data attribution framework with Shapley value \cite{2024efficientshapleyprune} and propose a method to approximate contribution of each training data point in the training process.
    \item \textbf{Copyright Loss Estimation:} 
    % to quantify the copyright loss of each training sample to compensate for the copyright infringement that it may have in the training process, we propose a novel metric of copyright loss to quantify the copyright infringement in subsection \nameref{subsec: copyright}.
    The copyright loss is a crucial consideration for data holders participating in model training. It is essential for the model holder to provide compensation to the data holder for this copyright loss. However, the absence of a robust metric that aligns closely with copyright law complicates the process of accurately estimating the copyright loss associated with training data
    % \item \textbf{Dynamic computation:} considering that the data in the training dataset will be added dynamically, the computational cost of calculating the data holders' contribution and copyright loss using traditional method, such as Monte Carlo method, will be very high. Therefore, we need to propose a time-efficient dynamic algorithm for contribution and copyright loss computation under the dynamic data setting.
    \item \textbf{Hierarchical Budget Allocation:} 
    The total budget $B$ will be distributed in a hierarchical structure. It should first be distributed to the training of each round, then to each data holder $k$ in the model training. It's difficult to decide an allocation that takes both the model performance and the compensation effect into account.
    % we propose a novel incentive system using RL to distribute the budget among different rounds and among the data holders that participated in the training. The RL based system distributes the budget among the rounds, and the copyright loss and contribution metric designs the payoff of each data holders in the training of the round.
\end{enumerate}

% \subsection{System Workflow}
% As shown in figure \ref{fig:workflow}, the workflow of our system can be described as follows:

% \begin{enumerate}
%     \item At round t, the agent of the outer layer RL allocates the budget $B_t$ to the inner layer.
%     \item Compute the contribution of each training sample $\hat{\beta}^t_i$ and the copyright loss of each training sample $c^t_i$.
%     \item Compute the reward $r$ of the inner layer RL.
%     \item The agent of the inner layer RL allocates the budget $p^t_i$ to each data holders by a new portion.
%     \item The inner layer RL finishes training and output the final selection of the data holders joining the training.
%     \item The model is trained with the selected data holders at the round t.
%     \item Repeat until the final round is reached, and compute the FID of the final model.
%     \item Repeat the above process until the stopping step of the outer RL is reached. Output the final model $M_T$ and the budget allocation.
% \end{enumerate}

\section{Main Approach}
In this section, we firstly model the contribution and copyright loss. Then we present a hierarchical incentive scheme based on contribution and copyright loss.
\subsection{Modeling Contribution}
\label{subsec:contribution}
To assess the contribution of each data holder in round $t$ within the diffusion model, we can simplify the problem to analyzing the contribution of a single training sample $i$. This involves determining how the distribution of generated output images is affected when a specific training sample is removed.

We consider the TRAK algorithm mentioned in Definition. \ref{def: trak}, and we need to decide the $\mathcal{L}$ to perform data attribution for the model. It's natural to consider the training objective as the $\mathcal{L}$:
% \begin{equation}
%     \mathcal{L}_{basic}= \mathbb{E}_{\epsilon,v}\norm{\epsilon-\epsilon_{\theta}(\sqrt{\overline{a_v}}x\,+\sqrt{1-\overline{a_v}}\epsilon,v)}^{2}_2.
% \end{equation}
\begin{equation}
    \mathcal{L}_{TIDM}=\mathbb{E}_{\epsilon,v}\norm{\epsilon-\epsilon_{\theta}(z_v,v, \mathcal{E}(l))}^{2}_2.
\end{equation}
However, as mentioned in follow-up algorithm D-TRAK \cite{zheng2024dtrak}, using a design choice of $\mathcal{L}$ can lead to a 40\% better data attribution performance than the default $\mathcal{L}:=\mathcal{L}_{TIDM}$, due to the reason that the default $\mathcal{L}$ is not the best at retaining the information of $\nabla_{\theta}\epsilon_{\theta}$. The revised $\mathcal{L}$ is formulated as follows:
\begin{equation}
    \mathcal{L}_{D-TRAK} := \mathbb{E}_{\epsilon,v}\norm{\epsilon_{\theta}(x_v,v,\mathcal{E}(l))}^{2}_2.
\end{equation}

In this case, the data attribution score will be formulated as:
\begin{equation}
    \tau_{\text{D-TRAK}}(x) = \frac{1}{N} \sum_{s=1}^{N} \hat{\phi}^s(x)^\top \left( \left( \Phi_{\text{D-TRAK}}^s \right)^\top \Phi_{\text{D-TRAK}}^s \right)^{-1} \left( \Phi_{\text{D-TRAK}}^s \right)^\top
\end{equation}

 Compared to the original TRAK formula \ref{eq: trak}, $\phi^s(x) = \mathbf{P}_s^\top \nabla_{\theta} \mathcal{F}(x; \theta_s^*)$ is changed to $\hat{\phi}^s(x)=\mathbf{P}_s^\top \nabla_{\theta}\mathcal{L}_{D-TRAK}$, and correspondingly $\Phi^s$ is changed to $\Phi^s_{\text{D-TRAK}} = \left[ \hat{\phi}^s(x^1); \cdots ; \hat{\phi}^s(x^K) \right]^\top$; $\mathcal{Q}^s$ is reduced to the identity matrix $I$ since we assume $\mathcal{L}$ and $\mathcal{F}$ are the same \cite{zheng2024dtrak}.

We use the above data attribution method in the text-to-image diffusion model and compute the score of each training sample as their contribution. The contribution of sample $i$ is formulated as:
\begin{equation}
    \beta_i=\tau_{D-TRAK}(x_i)
\end{equation}
We denote the contribution of training sample $i$ as $\beta^t_i$, the minimum and maximum of the contribution in the current round $t$ as $\beta^t_{min}$ and $\beta^t_{max}$, respectively, then, we normalize the raw contribution of each training sample as:
\begin{equation}
    \hat{\beta^t_i} = \frac{\beta^t_i - \beta^t_{min}}{\beta^t_{max}-\beta^t_{min}}
\end{equation}

The contribution of each data holder $k$ is computed as:
\begin{equation}
    \beta^t_k = \sum_{i} \hat{\beta^t_i}
\end{equation}

\subsection{Modeling Copyright Loss}
\label{subsec:copyright} 

\begin{figure}[htbp]
    \centering
    \includegraphics[width=\columnwidth]{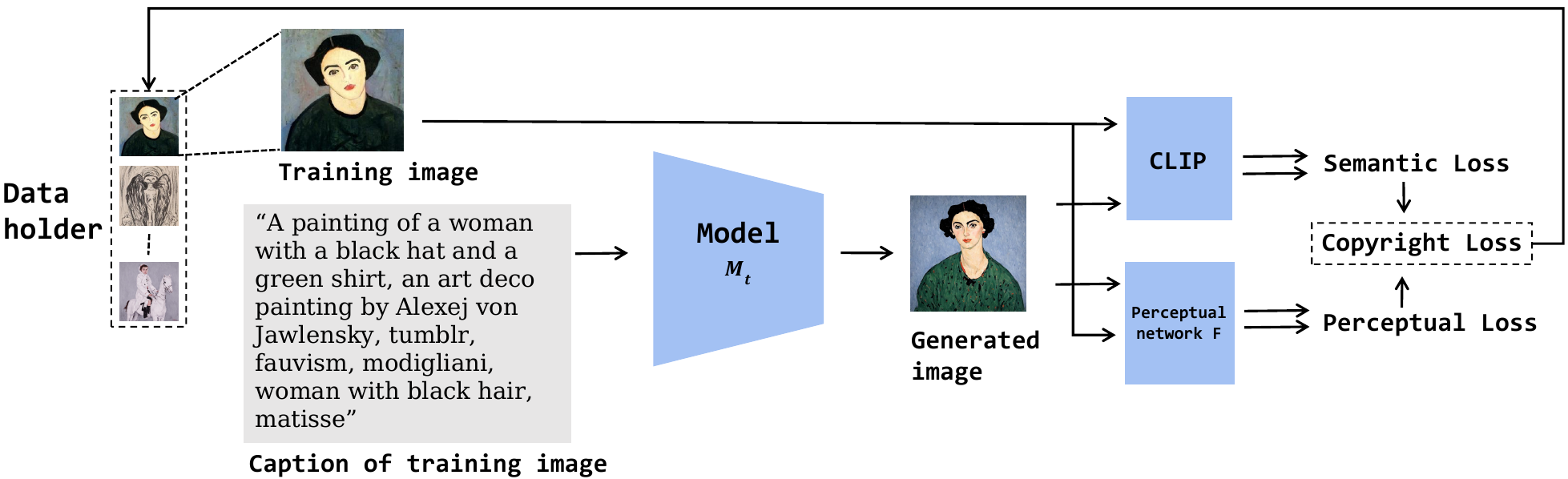}
    \vspace{-0.25in}
    \caption{Workflow of Computation of Copyright Loss.}
    \label{fig:copyright_workflow}
\end{figure}

Estimating copyright loss presents a significant challenge due to the difficulty in quantifying copyright infringement. While some previous studies have addressed copyright-related issues, they often lack explicit definitions of copyright infringement, instead using vague terms such as replications \cite{somepalli2022diffusion} and memorization \cite{carlini2023extracting}. For instance, \cite{vyas2023provable} introduces a stable algorithm as a surrogate for addressing copyright concerns, but it falls short in tracking copyright loss at the instance level for each training sample. Our goal is to propose a metric that accurately quantifies copyright infringement and assesses copyright loss associated with each individual sample.

According to US law, courts consider four factors when assessing fair use defenses \cite{Copyrightlaw1989}: (1) the purpose of the challenged use, including whether the use is of a commercial nature or is for nonprofit educational purposes, (2) the nature of the copyrighted works, (3) the amount and substantiality of the taking, and (4) the effect of the challenged use on the market for or value of the copyrighted work. From technical aspect, the 
substantiality will be the most measurable and relative determinants. The induced question will be: how to measure substantiality? In practice, the court will perform extrinsic and intrinsic tests to measure the substantiality \cite{Sid&Marty_1977}. The extrinsic test is an objective comparison of the expressive elements in the works. Then, intuitively, we need a metric that measures the similarity of the elements between original images and generated images. The intrinsic test is generally conducted by the jury, and it's related to the vision perception of human being and subjective feelings. This inspires us to propose metrics to measure substantiality from two aspects: (1) semantic metric and (2) perceptual metric.

\begin{figure*}[t]
    \centering
    \includegraphics[width=0.95\textwidth]{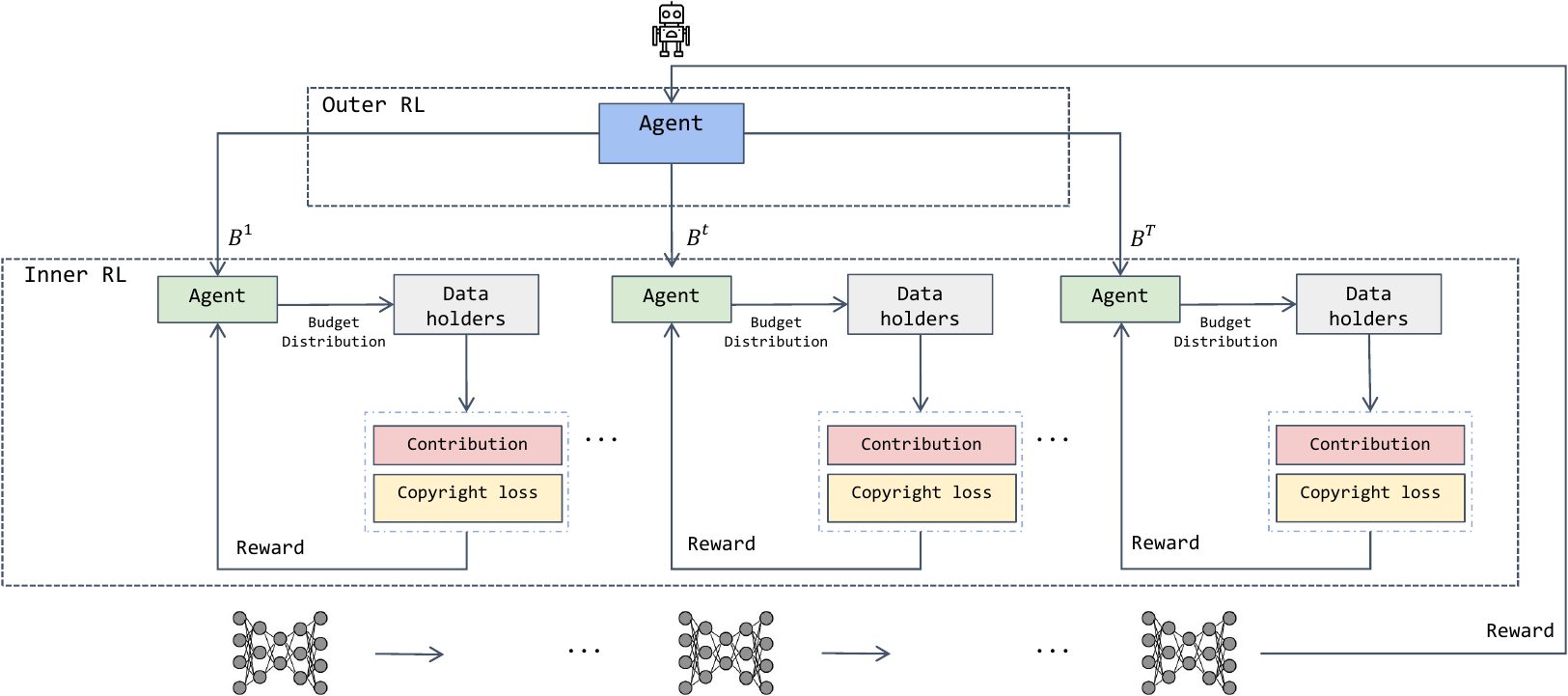}
    \vspace{-0.15in}
    \caption{Workflow of Hierarchical Reinforcement Learning. The outer RL distributes the budget $B^1, B^2, ..., B^T$ to each round. The inner RL distributes the budget to each data holder according to the contribution and the copyright loss, and the model with be trained with the corresponding data. The quality of the output model serves as the reward of the outer RL.}
    \label{fig:workflow}
\end{figure*}

\paragraph{Semantic Distance} The semantic metric measures the similarity of the extrinsic elements between the original image and the generated image. It is inspired by the extrinsic test in the court, and it tries to capture the similarity between the original image and the generated one. Following \cite{ma2024dataset}, we use CLIP \cite{radford2021learning} to generate the semantic embedding for each training sample and the generated sample, and compute the MSE for measurement. Denote the generated semantic embedding of training sample $s_i$ and the generated image by $CLIP(s_i)$ and $CLIP(y_i)$, respectively, then, 
\begin{equation}
\label{eq:semantic}
Dist_{sem} = MSE(CLIP(s_i)-CLIP(y_i))
\end{equation}

\paragraph{Perceptual Distance} We choose the perceptual metric as one of the measurement because it tries to mimic the vision perception of human being, and it aligns with the "the subjective comparison" by "reasonable audience" in the intrinsic test of the courts \cite{Sid&Marty_1977}. Following the framework of \cite{zhang2018unreasonable}, we can compute the perceptual similarity of the training sample and the generated image.
We compute deep embeddings with a network $\mathbf{F}$, then normalize the activations in the dimension of channel. Next, we scale the features with a weight and compute the average across the $l_2$ distance in the spatial dimensions, considering all layers.
\begin{equation}
    Dist_{per} = \sum_{l}\frac{1}{H_{l}W_{l}}\sum_{h,w}||w_{l}\odot(\hat{z_i}-\hat{y_i})||_2^2,
\end{equation}
where $\hat{z_i}$ and $\hat{y_i}$ are the deep embeddings of the training sample and the generated sample got from a deep network, $w_l$ is the scaled vector at layer $l$, $H_l$ and $W_l$ are the height and width of the feature map at layer $l$.

Then, the comprehensive similarity of sample $i$ is defined as: 
\begin{equation}
    c_i = a Dist_{sem} + b Dist_{per}
\end{equation}
where $a$ and $b$ are hyperparameters that control the importance attached to each measurement. 

We normalize and process the comprehensive similarity of each sample. The copyright loss of a sample $i$ at time $t$ is defined as follows:
\begin{equation}
    \hat{c_i}^t=1-\frac{c_i^t-c_{min}^t}{c_{max}^t-c_{min}^t}.
\end{equation}

\textbf{Workflow} Now we can compute the copyright loss with the metrics defined above for each training sample with respect to each generated image. As shown in Figure. \ref{fig:copyright_workflow}, the workflow for this part is as follows: initially, each training sample is passed through the CLIP model \cite{radford2021learning} to generate captions, which are fed into a latent diffusion model to generate corresponding images. Next, we calculate the semantic distance $Dist_{sem}$ using Equation \ref{eq:semantic}. Following this, in line with the framework proposed by Zhang et al. \cite{zhang2018unreasonable}, we compute the perceptual distance $Dist_{per}$. Once we have obtained both $Dist_{sem}$ and $Dist_{per}$, we proceed to calculate the comprehensive similarity $\hat{c_i}^t$ for each sample $i$ at iteration round $t$. Finally, the total copyright loss for data holder $k$ at round $t$ is determined by summing the individual scores across all samples, as defined in the equation below:
\begin{equation} 
c^t_k = \sum_i \hat{c^t_i} 
\end{equation}

\newlength\myindent
\setlength\myindent{2em}
\newcommand\bindent{%
    \begingroup
    \setlength{\itemindent}{\myindent}
    \addtolength{\algorithmicindent}{\myindent}
}
\newcommand\eindent{\endgroup}

\begin{algorithm}[b!]
\small
\caption{Hierarchical Budget Allocation}
\label{alg:budget_allocation}

\begin{algorithmic}[1]
\item[] \textbf{Input:} Total budget $B$; pre-trained model $\mathcal{M}$; the data holders $\mathcal{D}_1, \mathcal{D}_2, ..., \mathcal{D}_n $; round number $T$.
\item[] \textbf{Output:} Trained model $\mathcal{M}_T$, the budget allocation at each round $B^1, B^2, ..., B^T$.

\item[] \textbf{Outer Budget Allocation:}
\State Initialize Q-network with random weights
\State Initialize replay buffer $\mathcal{D}$
\For{each time slot $t = 1$ to $T$}
    % \Indent
    \State Observe state $\mathcal{S}_t = (N_t, C_t, X_t, B_{lt}, t_l)$
    \State \parbox[t]{\dimexpr\linewidth-\algorithmicindent}{Select action $\mathcal{A}_t = B^T$ based on $\epsilon$-greedy policy from Q-network}
        
        \State \parbox[t]{\dimexpr\linewidth-\algorithmicindent}{\textbf{Inner Budget Allocation:}}
        % \State Initialize Q-network for inner allocation with random weights
        \State \parbox[t]{\dimexpr\linewidth-\algorithmicindent}{Initialize Q-network for inner allocation with random weights}

        \State Initialize replay buffer $\mathcal{D}_{inner}$
        \If{$t = 1$}
            \State Load the pre-trained model $\mathcal{M}$
        \EndIf
        \For{iteration $j = 1$ to $10^5$}
            \State Compute the contribution and copyright of each images $i$
            \State \parbox[t]{\dimexpr\linewidth-\algorithmicindent}{Select action $\mathbf{p}^t = (p^t_1, p^t_2, \ldots, p^t_n)$, where each $p^t_k$ represents the fraction of the budget distributed to data holder $k$}
            
            \State Distribute budget: $P^t = B^t \times \mathbf{p}^t$
            \State \parbox[t]{\dimexpr\linewidth-\algorithmicindent}{Compute rewards: $r_t = \lambda R_{con} - \delta R_{cop}$, where: $R_{con} = \sum_k {p_k^t} \times \hat{\beta_k}^t, \quad R_{cop} = \sum_k {p_k^t} \times \hat{c_k}^t$}
            \State Store transition $(B^t, \mathbf{p}^t, r_t)$ in $\mathcal{D}_{inner}$
            \State Sample mini-batch from $\mathcal{D}_{inner}$
            \State Update Q-network for inner allocation

        \EndFor
        \State Train the model with selected data
    \State Receive outer RL reward $\mathcal{R}_t$
    \If{$t_l > 1$}
        \State $\mathcal{R}_t = 0.01$
    \Else
        \State $\mathcal{R}_t = Q(\mathcal{M})$ \Comment{Model quality measured by FID score}
    \EndIf
    \State Store transition $(\mathcal{S}_t, \mathcal{A}_t, \mathcal{R}_t, \mathcal{S}_{t+1})$ in $\mathcal{D}$
    \State Sample mini-batch from $\mathcal{D}$
    \State Update Q-network
    % \EndIndent
\EndFor

% \item[] \textbf{Return:} Trained model $\mathcal{M_T}$

\end{algorithmic}
\end{algorithm}

\subsection{Hierarchical Budget Allocation}
The budget allocation problem could be divided into distributing the budget across time slots and within a specific time slot among data holders. To deal with this issue, we use the hierarchical RL as the backbone framework as it leverages the adaptive nature of RL and is able to handle the complexity in different stages \cite{hierarchical_rl}. 

\subsubsection{Outer Budget Allocation}
The outer budget allocation in the hierarchical budget allocation method is used to allocation the total budget $B$ among time slots $t$, i..e,  training rounds of the model. It is developed based on the DQN framework \cite{li2017deep} and detailed as:

\textbf{State}: The state of the outer budget allocation is composed of 1) historical information of the previous rounds, including the amount of data points contributed by participating data holders $N_t$, the total copyright loss in the previous training up to time $t$, denoted as $C_t$, the total contribution in the previous training up to time $t$, denoted as $X_t$ 2) the leftover budget till time slot $t$, denoted as $B_{lt}$, and 3) the maximum leftover time slot $t_l$:
\begin{equation}
    \mathcal{S}_t = (N_{t}, C_{t}, X_{t}, B_{l,t}, t_l)
\end{equation}

\textbf{Action}: The action of the outer budget allocation is the budget for the current round, denoted as $\mathcal{A}_t$. 

\textbf{Reward}: The reward $\mathcal{R}_t$ is formulated as the quality of the generative model measured by the Fréchet Inception Distance (FID) score \cite{fid2018}:
\begin{equation}
\mathcal{R}_t = \left\{
\begin{array}{rcl}
0.01  &    & {if t_l > 1}\\
Q(\mathcal{M})               &    & {otherwise}
\end{array} \right.
\end{equation}
The $Q(\mathcal{M})$ is the output quality of the model (See Equation. \ref{eq:model_quality}). The design of the reward $R_t$ means that at the last round $T$, the reward will be the model quality measured by the metric \ref{eq:model_quality}; at the middle rounds $t<T$, we will attribute a small amount of reward to encourage the budget distribution process.

% We model the Reinforcement Learning process as a Markov Decision Process (MDP), which consists of a state set $\mathcal{S}$, an action set $\mathcal{A}$ and a reward function $\mathcal{R} : \mathcal{S} \times \mathcal{A} \longrightarrow \mathbb{R}$.

% At time slot $t$, the parameters of the MDP is denoted as:

% \[, \mathcal{A}_t = B_t\]

% \begin{math}
% \mathcal{R}_t = \left\{
% \begin{array}{rcl}
% B_{l,t}  &    & {if t_l = 1}\\
% Q(\mathcal{M})               &    & {otherwise}
% \end{array} \right.
% \end{math}

% where $N_{t}$ is the total number of the data holders joined in the training before time t, $C_{t}$ is the total copyright loss in the previous training up to time t, $D_{t}$ is the total contribution in the previous training up to time t, $B_{lt}$ is the leftover budget for time slot $t$ and future time slots, $t_l$ denotes the maximum leftover time slots i.e. $T - t$, and $Q(\mathcal{M})$ is the quality of the generated image of the model in the last time step. We use FID as a metric to evaluate the quality of the generated images of the generative art model.

\subsubsection{Inner Budget Allocation}
The inner budget allocation aims to allocate the round budget $B^T$, i.e., $\mathcal{A}_t$ described above, among the various data holders participating in the corresponding training round. Specifically, similar to the outer budget allocation, the inner budget allocation is also grounded in a DQN framework:
% In a specific time slot $t$, we distribute the payoff $P^t$ to different data holders by another RL system. We use a DQN framework and the settings are as described:

\textbf{State}: The state of the inner budget allocation is round budget $B^T$, i.e., the action of the outer budget allocation.

\textbf{Action}: The action of the inner budget allocation is denoted as a vector $p^t$, recording the fraction of the budget distributed to different data holders at the time slot $t$.
\begin{equation}
    \mathbf{p^t} = \begin{pmatrix}
p^t_1 & p^t_2 & \cdots & p^t_k & \cdots & p^t_n
\end{pmatrix},
\end{equation}
where $p_1 + p_2 + ... + p_n = 1$. Then, the payment for each data holder $k$ in round $t$, denoted as $P^t_k$ can be computed by the fraction:
\begin{equation}
    P^t_k = B^t \times p^t_{k}
\end{equation}

\textbf{Reward}: The reward is formulated as:
\begin{equation}
    \label{eq:inner_reward}
    r_t = \lambda R_{con} - \delta R_{cop},
\end{equation}
where
% $R_{con} = \sum_k {p_k} \times \hat{\beta_k}$ and
% $R_{cop} = \sum_k {p_k} \times c_k$. 
$R_{con} = \sum_k {p_k^t} \times \hat{\beta_k}^t, \quad R_{cop} = \sum_k {p_k^t} \times \hat{c_k}^t$
The reward is based on the intuition that data holder with high contribution and high copyright should be given high payoffs. We design the reward to help the system pick the data holders with the larger contribution and the lower the copyright loss. The setting can help the system spend the budget in a smart way that allocate more payoff to the data holders that make greater contribution to the training while doesn't need to compensate a lot for the copyright loss.

\subsubsection{Workflow}
Figure \ref{fig:workflow} shows the workflow of the Hierarchical Budget Allocation proposed in our method, which is described as:
\begin{enumerate}
    \item At round $t$, the agent of the outer layer RL allocates the budget $B^t$ to the inner layer.
    \item Train the model with data in all data holders. Compute the contribution of each training sample $\hat{\beta}^t_i$ and the copyright loss of each training sample $\hat{c_i}^t$.
    \item The agent of the inner RL get the reward $r$ of the inner layer RL according to Equation. \eqref{eq:inner_reward}.
    \item The agent of the inner layer RL allocates the budget $p^t_i$ to each data holders with a decided portion $p_t$.
    \item The agent of the inner RL get a new reward $r$ of the inner layer RL according to Equation. \eqref{eq:inner_reward}, and repeat the process until the stopping step.
    \item The inner layer RL finishes training and output the final selection of the data holders joining the training.
    \item The model is trained with the selected data holders at the round t.
    \item Repeat until the final round is reached and compute the model quality  of the final model according to Equation. \eqref{eq:model_quality}. The value will be the reward of the outer layer RL.
    \item Repeat the above process until the stopping step of the outer RL is reached. Output the final model $M_T$ and the budget allocation.
\end{enumerate}

The proposed Hierarchical Budget Allocation method is
summarized in Algorithm. \ref{alg:budget_allocation}.

\begin{table*}[!t]
\centering
\caption{Model Output Quality for Comparative Experiments on Three Datasets}
\vspace{-0.15in}
\label{table1}
\resizebox{0.95\textwidth}{!}{
\begin{tabular}{|c|*{9}{c|}}
\hline
Dataset & Ours & G+L & L+L & R+R & RL+R & RL+L & R+RL & L+RL & G+RL \\ 
\hline
ArtBench &  \textbf{6.178$\pm$0.344} &  5.065$\pm$0.161&  3.568$\pm$0.020&  3.842$\pm$0.366&  5.692$\pm$0.470&  5.782$\pm$0.114&  5.017$\pm$0.609&  4.259$\pm$0.122&  5.993$\pm$0.279\\ 
\hline
Portrait &  \textbf{8.331$\pm$1.785}&  5.767$\pm$0.106&  5.986$\pm$0.081&  6.017$\pm$0.966&  8.169$\pm$2.011&  8.186$\pm$1.257&  6.057$\pm$1.397&  7.509$\pm$0.425&  5.943$\pm$0.195\\ 
\hline
Cartoon & \textbf{6.038$\pm$0.737}
& 5.032$\pm$0.644& 3.955$\pm$0.211& 4.508$\pm$1.307& 5.613$\pm$0.707& 5.947$\pm$0.520& 4.594$\pm$0.973& 4.219$\pm$0.413& 4.516$\pm$0.529  \\ 
\hline
\end{tabular}}
\end{table*}
\begin{figure*}[t] % 't' ensures it's placed at the top of the page across both columns
    \centering
    \begin{subfigure}[b]{0.33\textwidth}  % Adjust width for three images
        \includegraphics[width=\textwidth]{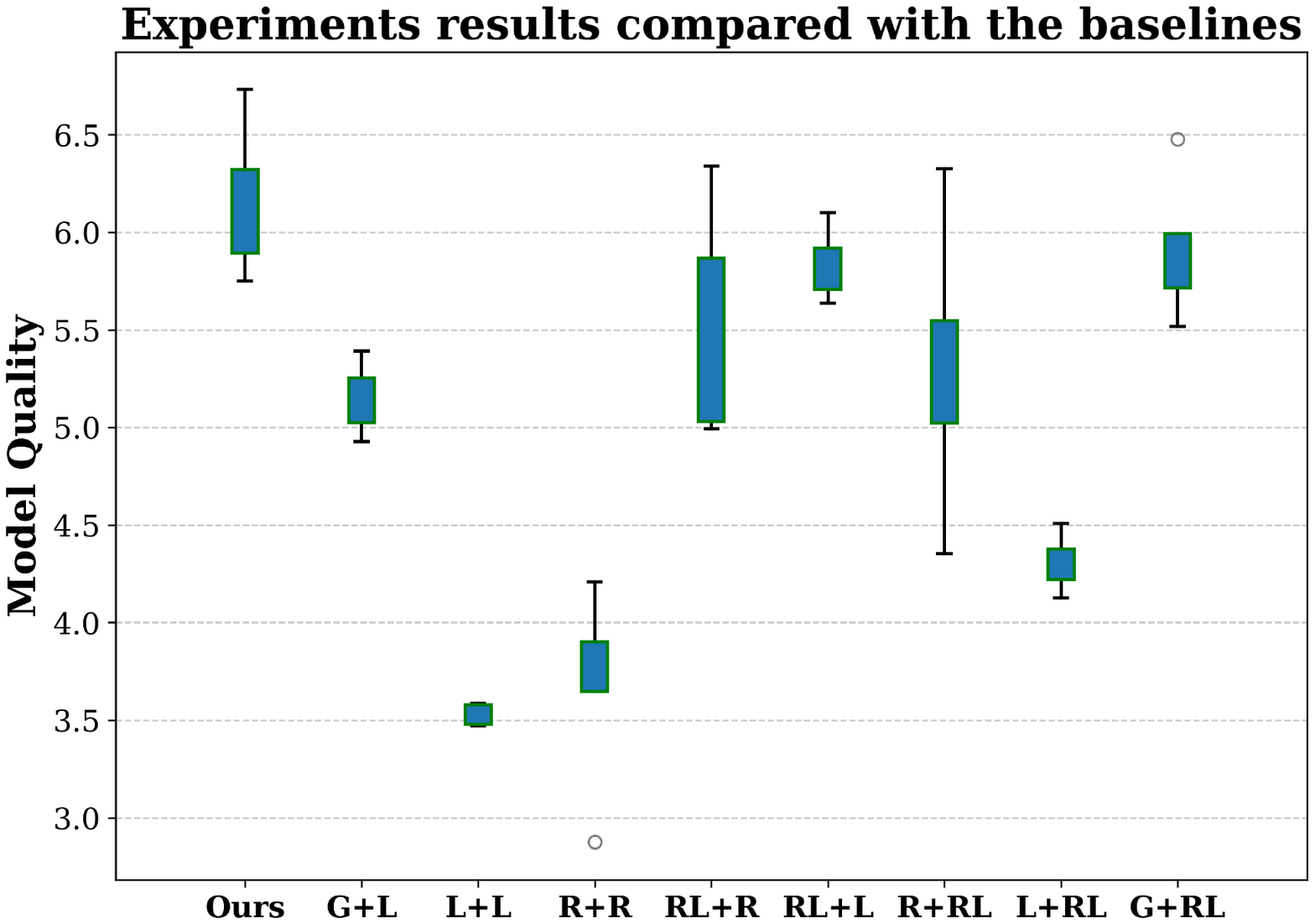}
        \vspace{-0.2in}
        \caption{ArtBench}
        \label{fig:result1}
    \end{subfigure}
    \hfill  % Space between the subfigures
    \begin{subfigure}[b]{0.33\textwidth}
        \includegraphics[width=\textwidth]{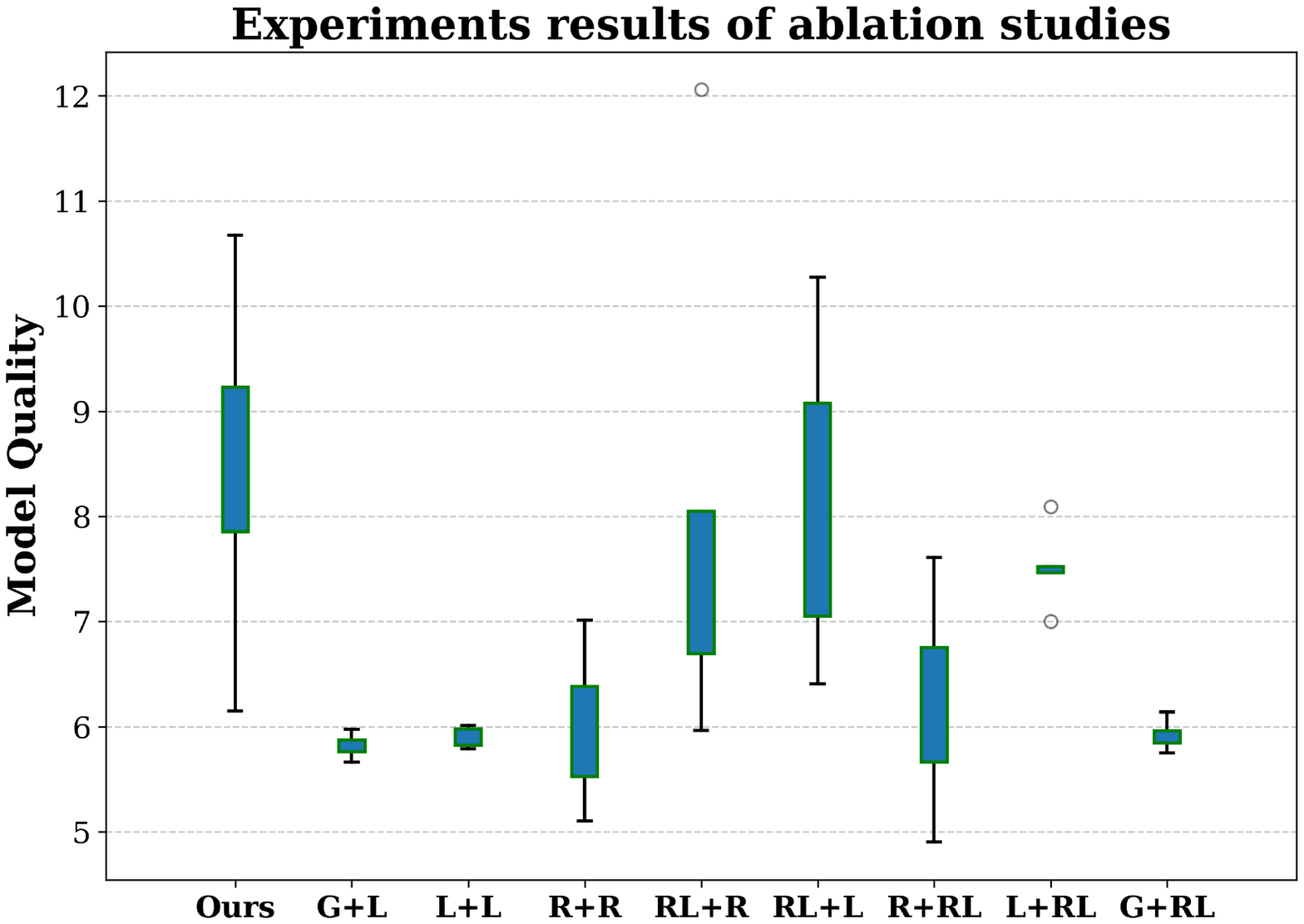}
         \vspace{-0.2in}
        \caption{Portrait}
        \label{fig:result2}
    \end{subfigure}
    \hfill
    \begin{subfigure}[b]{0.33\textwidth}
        \includegraphics[width=\textwidth]{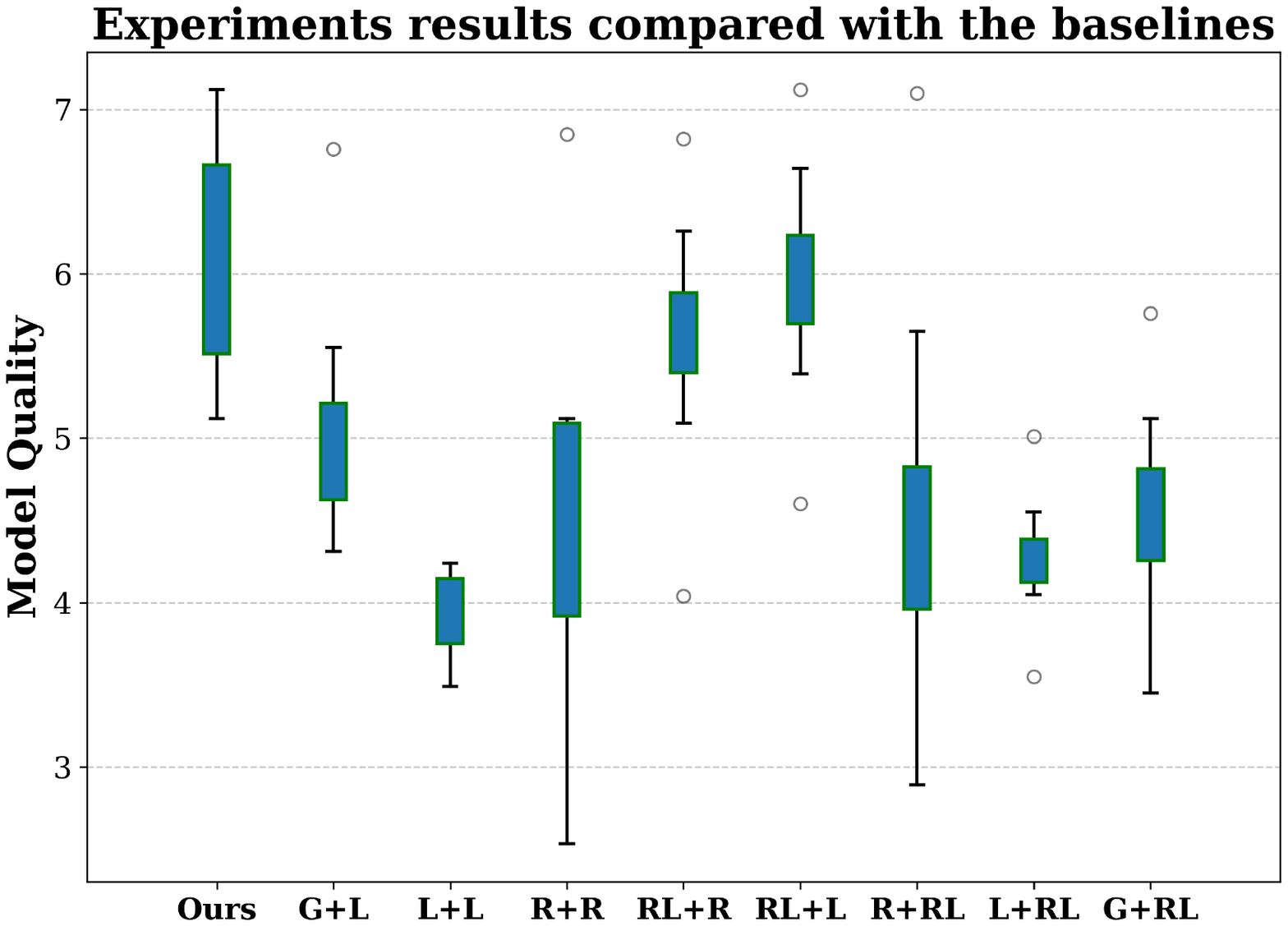}
         \vspace{-0.2in}
        \caption{Cartoon}
        \label{fig:result3}
    \end{subfigure}
    \vspace{-0.3in}
    \caption{Comparison of Our Method with the Baselines on Three Datasets.}
    \label{fig:main_results}
\end{figure*}
\section{Experiments}
\subsection{Experiment Setup}
\textbf{Dataset. } We use the following three datasets:
\begin{enumerate}
    \item \textbf{ArtBench: } ArtBench\cite{liao2022artbench} includes 60000 selected high-quality artworks in 10 distinctive art styles with 256x256 resolution. We gathered 800 256x256 images from the Impressionism subset.
    \item \textbf{Portrait: } The Portrait dataset\cite{ma2024dataset} includes 200 images of celebrities from Wikipedia with detailed annotations about the outfit, facial features and gesture of the people. The copyright of a portrait encompasses an individual’s authority over their own image, including their facial features, likeness, and posture.
    \item \textbf{Cartoon: } The cartoon dataset\cite{ma2024dataset} includes 200 images of cartoon characters from animations and cartoons with detailed notations. The images are selected from Wikipedia\footnote{\url{https://www.tensorflow.org/datasets/catalog/wikipedia}}. These cartoon images, featuring unique graphic expressions, are often considered copyrighted due to their distinctive characteristics.
\end{enumerate}

\textbf{Environments and models. } We conducted the experiments on 4 NVIDIA A-100. The diffusion model we used is one of the newest models in Stable Diffusion series: Stable Diffusion XL 1.0\footnote{\url{https://github.com/huggingface/diffusers}}. The inner and outer layer of hierarchical RL are modified from the DQN framework based on the opensource repository ElegantRL\footnote{\url{https://github.com/AI4Finance-Foundation/ElegantRL}}. The explore rate of the DQN network is 0.5, and the discount factor of the future rewards is 0.98. The middle layer dimension of MLP is set to [80, 40]. The weight for both semantic similarity and perceptual similarity is set to 0.5, i.e., $a = b = 0.5$. $\lambda$ and $\delta$ in the reward function of the inner budget allocation is set to 0.5.

\textbf{Implementation Settings}. We have 8 data holders with different amount (50, 60, 80, 100, 150, 200)  and different quality (low, medium, high) of training data in the system. Each data holder will have a self-proposed price for their data, which is publicly known. When the allocated payment meet the self-proposed price, the data holder will join the training for the current round.

\textbf{Evaluation Metrics. } We use a modified version of Fréchet Inception Distance (FID) score \cite{fid2018} $Q(\mathcal{M})$ to measure the output quality of diffusion models:
\begin{equation}
    \label{eq:model_quality}
    Q(\mathcal{M}) := \frac{100}{FID+10^{-6}}. 
\end{equation}
FID measures how similar the distribution of generated images is to the distribution of real images, based on features extracted by a pre-trained Inception network. The formulation of the metric is as follows:
\begin{equation}
\text{FID} = \|m_x - m_y\|^2 + \text{Tr}(\text{Cov}(x) + \text{Cov}(y) - 2(\text{Cov}(x) \text{Cov}(y))^{1/2})
\end{equation}
where:
\(m_x\) and \(\text{Cov}(x)\) represent the mean and covariance matrix of the feature vectors from real images;
\(m_y\) and \(\text{Cov}(y)\) represent the mean and covariance matrix of the feature vectors from generated images;
\(\text{Tr}(\cdot)\) denotes the trace of a matrix.

%% Outer + Inner
% 1. Greedy + Linearly
% 2. Linearly + Linearly
% 3. Random + Random
% 4. RL + Random
% 5. RL + Linearly
% 6. RL + contribution
% 7. RL + copyright loss
% 8. Random + RL
% 9. Linearly + RL
% 10. Greedy + RL
% 11. main (double rl)

\begin{figure*}[t] % 't' ensures it's placed at the top of the page across both columns
    \centering
    \begin{subfigure}[b]{0.33\textwidth}  % Adjust width for three images
        \includegraphics[width=\textwidth]{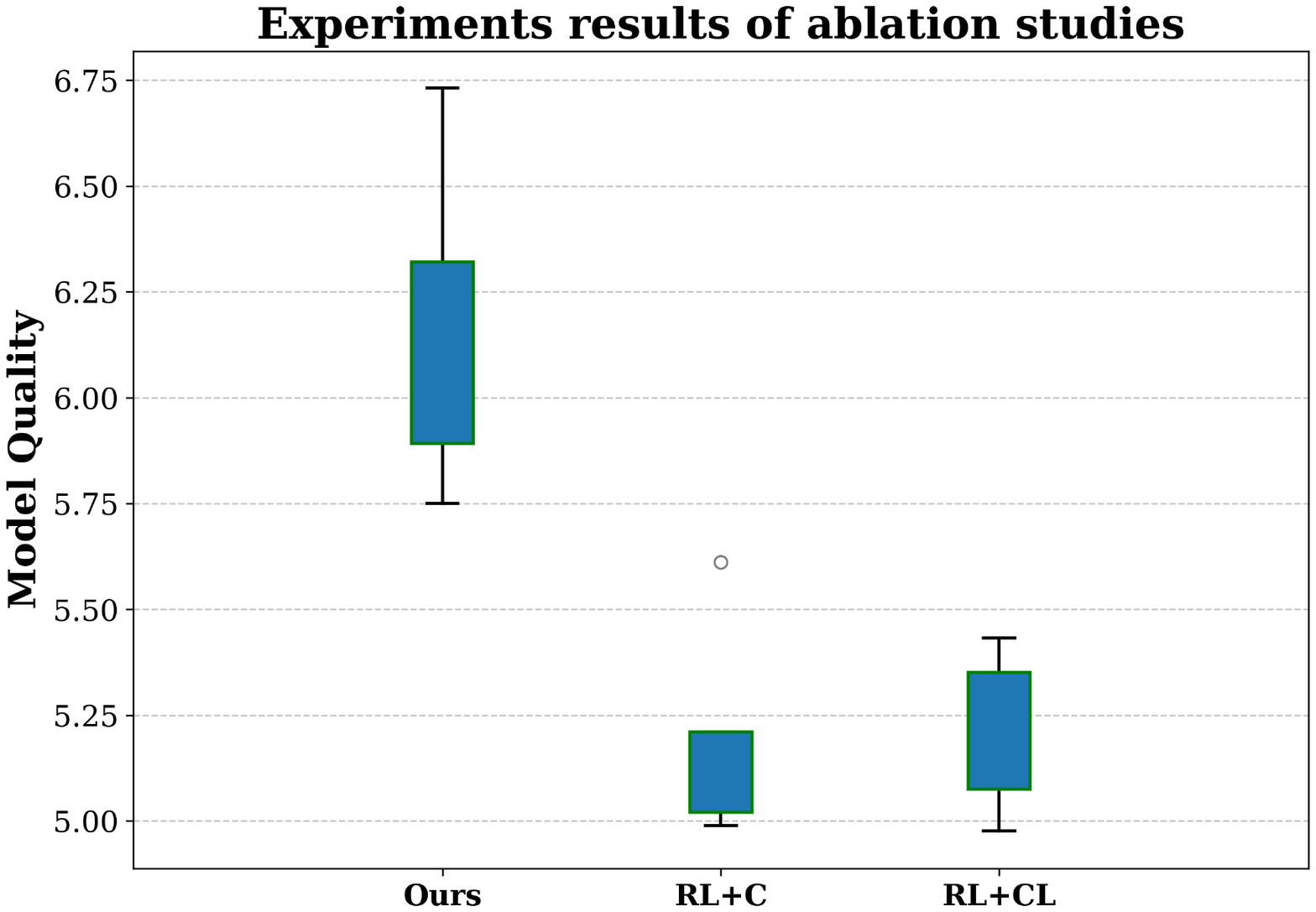}
         \vspace{-0.2in}
        \caption{ArtBench}
        \label{fig:ablation1}
    \end{subfigure}
    \hfill  % Space between the subfigures
    \begin{subfigure}[b]{0.33\textwidth}
        \includegraphics[width=\textwidth]{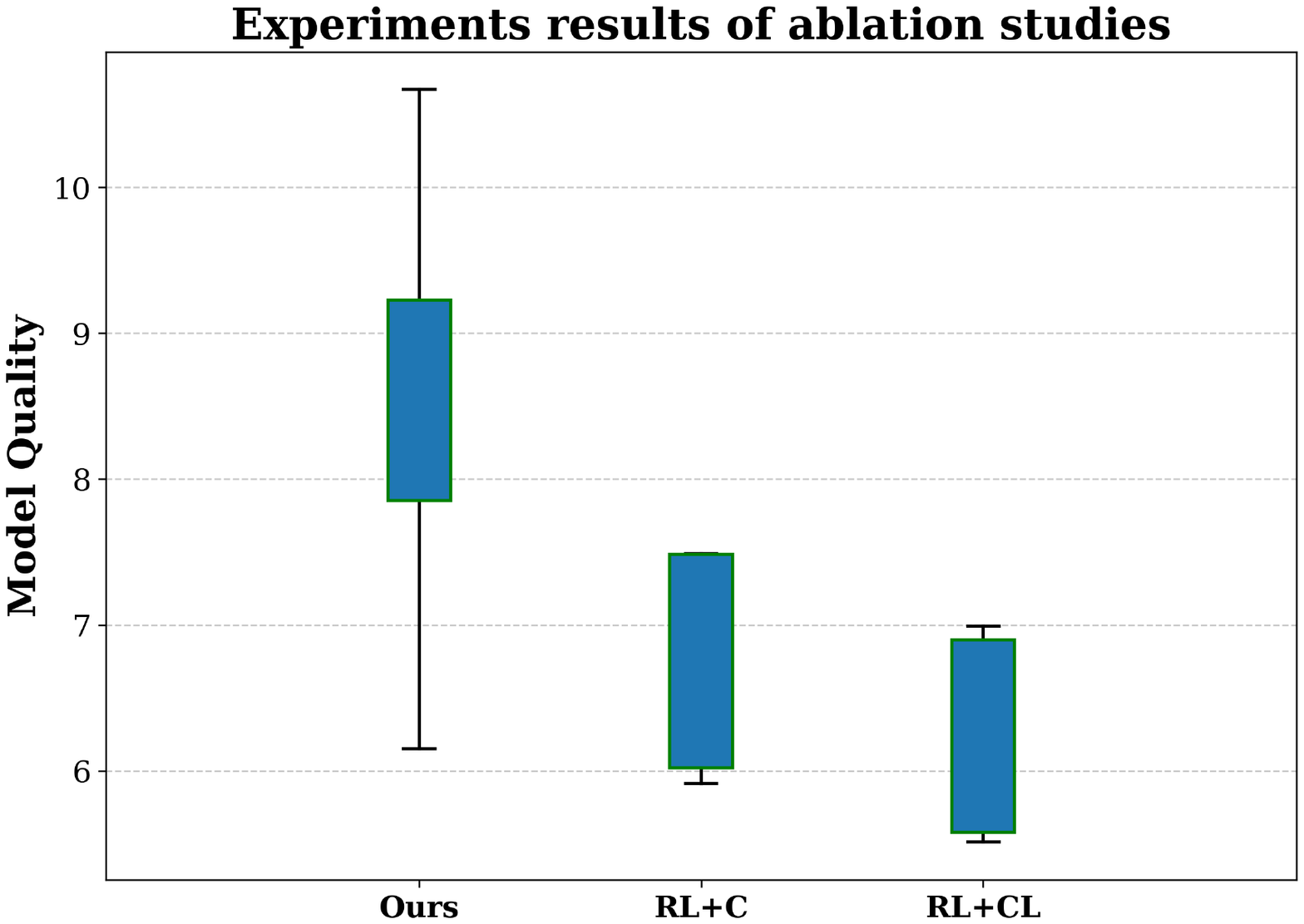}
        \vspace{-0.2in}
        \caption{Portrait}
        \label{fig:ablation2}
    \end{subfigure}
    \hfill
    \begin{subfigure}[b]{0.33\textwidth}
        \includegraphics[width=\textwidth]{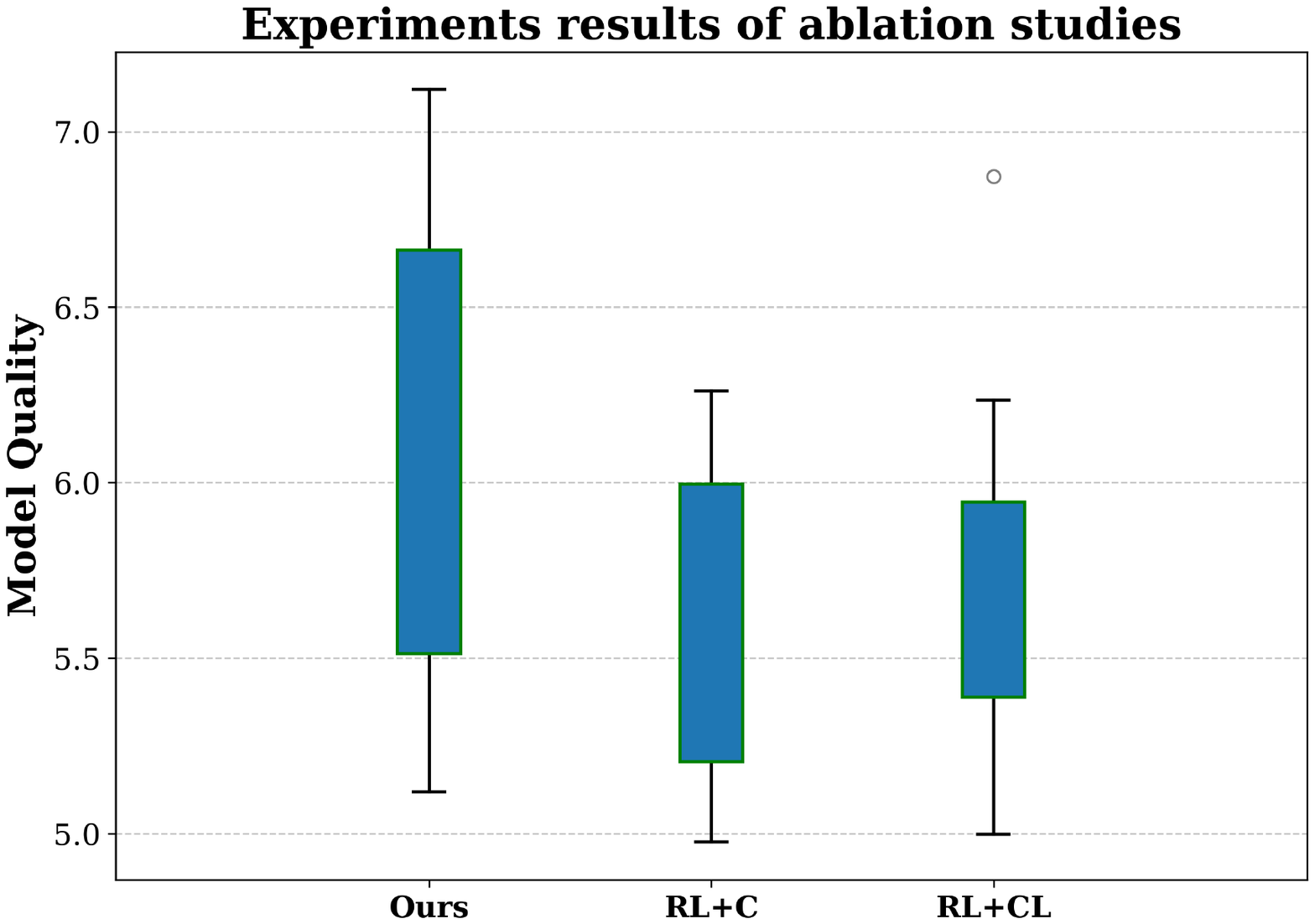}
        \vspace{-0.2in}
        \caption{Cartoon}
        \label{fig:ablation3}
    \end{subfigure}
    \vspace{-0.3in}
    \caption{Ablation Results on Three Datasets.}
\end{figure*}
\textbf{Baseline. } We integrate three strategies: random, linear, and greedy strategies to the experiments and compare the proposed method with 10 different baselines: 
\begin{enumerate}
    % \item \textbf{Random}: the method applies random budget distribution approach on TRAK score, copyright loss and the amount of the data in each dataholder.
    % \item \textbf{Linear}: the method applies RL for budget distribution with a linear distribution approach on TRAK score, copyright loss and the amount of the data in each dataholder.
    % \item \textbf{Greedy+linear}: 
    \item \textbf{Greedy + Linear(G+L)} uses greedy method for outer layer budget distribution, and the linear method for inner layer budget distribution;
    \item \textbf{Linear + Linear(L+L)} linearly distributes the budget at the outer and inner layer;
    \item \textbf{Random + Random(R+R)} randomly distributes the budget at the outer and inner layer;
    \item \textbf{RL + Random(RL+R)} uses RL for outer layer budget distribution, and randomly distributes the budget at the inner layer;
    \item \textbf{RL + Linear(RL+L)} uses RL for outer layer budget distribution, and linearly distributes the budget at the inner layer;
    \item \textbf{Random + RL(R+RL)} randomly distributes the budget at the outer layer, and uses the RL for inner distribution;
    \item \textbf{Linear + RL(L+RL)} linearly distributes the budget at the outer layer, and uses RL for inner distribution;
    \item \textbf{Greedy + RL(G+RL)} uses greedy method for outer layer budget distribution, and uses the RL for inner distribution.
\end{enumerate}
%  \textbf{Ablation Study}
% \begin{enumerate}
%     \item \textbf{RL + Contribution} uses the RL framework for outer distribution, and distributes the budget according to the contribution of the data holder at the inner layer;
%     \item \textbf{RL + Copyright loss} uses the RL framework for outer distribution, and distributes the budget according to the copyright loss of the data holder at the inner layer;
% \end{enumerate}

\subsection{Results and Discussion}
The comparision results are shown in Table \ref{table1} and Figure. \ref{fig:main_results}.
It can be observed that on all the three datasets, our method consistently outperform the baselines in terms of the model quality. Our proposed method can optimize the budget distribution and maximize the quality of the model, which validates the effectiveness of the proposed method.

We can also find that
\begin{enumerate}
    \item Methods using RL generally outperforms other methods in terms of the model quality. Most top-performing approaches integrate RL, as RL has strong abilities of finding optimal budget distributions. Our method with hierarchical RL structure can optimize the outcome from a global perspective and achieve the highest performance.
    \item Among the methods using single layer RL, methods using RL at the outer layer outperform the methods using RL at the inner layer in terms of model quality. The results imply that the global decisions that the outer RL deals with might have a larger impact on the overall model performance.
    \item Methods using linear distribution in both layers have bad performance in terms of model quality, since uniform allocation does not capture differences between data holders and between rounds.
    \item Methods combining random strategy are generally not well-performed. The random strategy does no optimization for the system, and it has a considerably large variance comparing to the other methods due to the randomness.
    % \item Methods combining greedy strategy outperforms those combining random or linear strategy. The greedy strategy focuses on making the best possible decision at each step, and this can lead to good short-term gain.
\end{enumerate}

\subsubsection{Ablation Study}

We conduct ablation experiments to explore the impact of only considering the contributions of data participants and copyright loss on the results. 

\begin{enumerate}
    \item \textbf{RL + Contribution(RL+C)} uses RL for outer distribution, and distributes the budget according to the contribution of the data holder at the inner layer;
    \item \textbf{RL + Copyright loss(RL+CL)} uses RL for outer distribution, and distributes the budget according to the copyright loss of the data holder at the inner layer;
\end{enumerate}

As shown in Table \ref{ablation}, our method outperforms both the ablated versions in terms of model quality. 
% The results show that combining the contribution and the copyright loss in the budget distribution is necessary and important. 
The results show that the contribution and copyright loss are both playing crucial roles in enhancing the model performance on efficiently allocate the budget.

\begin{table}[!t]
\caption{Ablation Studies on Three Datasets}\label{ablation}
\vspace{-0.15in}
\resizebox{0.45\textwidth}{!}{
\begin{tabular}{|c|*{3}{c|}}
\hline
Dataset & Ours & RL+C & RL+CL\\ 
\hline
ArtBench &\textbf{6.178$\pm$0.344}&  5.102$\pm$0.251&  5.284$\pm$0.121\\ 
\hline
Portrait &  \textbf{8.331$\pm$1.785}&  6.759$\pm$0.764& 6.390$\pm$0.695\\ 
\hline
Cartoon &\textbf{6.038$\pm$0.737}& 5.624$\pm$0.490& 5.741$\pm$0.616  \\ 
\hline
\end{tabular}}
\end{table}

% \begin{figure*}[t] % 't' ensures it's placed at the top of the page across both columns
%     \centering
%     \begin{subfigure}[b]{0.33\textwidth}  % Adjust width for three images
%         \includegraphics[width=\textwidth]{artbench_ablation(1).pdf}
%         \caption{ArtBench}
%         \label{fig:ablation1}
%     \end{subfigure}
%     \hfill  % Space between the subfigures
%     \begin{subfigure}[b]{0.33\textwidth}
%         \includegraphics[width=\textwidth]{ablation2.pdf}
%         \caption{Portrait}
%         \label{fig:ablation2}
%     \end{subfigure}
%     \hfill
%     \begin{subfigure}[b]{0.33\textwidth}
%         \includegraphics[width=\textwidth]{ablation3.pdf}
%         \caption{Cartoon}
%         \label{fig:ablation3}
%     \end{subfigure}
%     \caption{Ablation results on three datasets.}
% \end{figure*}

% \begin{figure}[h]
%     \centering
%     \includegraphics[width=\columnwidth]{ablation1.pdf}
%     \caption{Ablation Results.}
%     \label{fig:result1}
% \end{figure}

% \begin{figure}[h]
%     \centering
%     \includegraphics[width=\columnwidth]{result1.pdf}
%     \caption{The comparison of our method and the baselines on ArtBench.}
%     \label{fig:result1}
% \end{figure}

% \begin{figure}[h]
%     \centering
%     \includegraphics[width=\columnwidth]{results2.pdf}
%     \caption{The comparison of our method and the baselines on Cartoon.}
%     \label{fig:result2}
% \end{figure}

% \begin{figure}[h]
%     \centering
%     \includegraphics[width=\columnwidth]{results3.pdf}
%     \caption{The comparison of our method and the baselines on Portrait.}
%     \label{fig:result3}
% \end{figure}

\section{Conclusion and Future Work}

This paper introduces a novel method to address the challenge of copyright infringement in generative art with an economic method. Our approach introduces a novel copyright metric, grounded in legal perspective, to evaluate copyright loss and integrates the TRAK method to quantify the contribution of data holders. Through a hierarchical budget allocation method using reinforcement learning, we demonstrate how remuneration can be fairly distributed based on each data holder’s contribution and copyright impact. Extensive experiments validate the efficacy of this scheme, significantly outperforming existing incentive scheme in terms of model quality. 

% Our current approach doesn't consider the computing contribution and copyright loss in a dynamic situation, e.g. when the 
In the future, we plan to expedite the calculation of copyright loss and contributions to adapt the model for dynamic environments where data holders may join and leave. Additionally, current methods that assign identical copyright losses to duplicated samples can result in repeated payments for the same content. This issue calls for further exploration to refine the handling of duplicates in datasets.
\newpage

\bibliographystyle{ACM-Reference-Format}
\bibliography{main}


\begin{thebibliography}{45}

%%% the Web address.

\ifx \showCODEN    \undefined \def \showCODEN     #1{\unskip}     \fi
\ifx \showDOI      \undefined \def \showDOI       #1{#1}\fi
\ifx \showISBNx    \undefined \def \showISBNx     #1{\unskip}     \fi
\ifx \showISBNxiii \undefined \def \showISBNxiii  #1{\unskip}     \fi
\ifx \showISSN     \undefined \def \showISSN      #1{\unskip}     \fi
\ifx \showLCCN     \undefined \def \showLCCN      #1{\unskip}     \fi
\ifx \shownote     \undefined \def \shownote      #1{#1}          \fi
\ifx \showarticletitle \undefined \def \showarticletitle #1{#1}   \fi
\ifx \showURL      \undefined \def \showURL       {\relax}        \fi
% invisible to TeX
\providecommand\bibfield[2]{#2}
\providecommand\bibinfo[2]{#2}
\providecommand\natexlab[1]{#1}
\providecommand\showeprint[2][]{arXiv:#2}

\bibitem[Bourtoule et~al\mbox{.}(2020)]%
        {bourtoule2020machine}
\bibfield{author}{\bibinfo{person}{Lucas Bourtoule}, \bibinfo{person}{Varun Chandrasekaran}, \bibinfo{person}{Christopher~A. Choquette-Choo}, \bibinfo{person}{Hengrui Jia}, \bibinfo{person}{Adelin Travers}, \bibinfo{person}{Baiwu Zhang}, \bibinfo{person}{David Lie}, {and} \bibinfo{person}{Nicolas Papernot}.} \bibinfo{year}{2020}\natexlab{}.
\newblock \bibinfo{title}{Machine Unlearning}.
\newblock
\newblock
\showeprint[arxiv]{1912.03817}~[cs.CR]


\bibitem[Cao et~al\mbox{.}(2024)]%
        {generative_diffusion_model_survey}
\bibfield{author}{\bibinfo{person}{Hanqun Cao}, \bibinfo{person}{Cheng Tan}, \bibinfo{person}{Zhangyang Gao}, \bibinfo{person}{Yilun Xu}, \bibinfo{person}{Guangyong Chen}, \bibinfo{person}{Pheng-Ann Heng}, {and} \bibinfo{person}{Stan~Z. Li}.} \bibinfo{year}{2024}\natexlab{}.
\newblock \showarticletitle{A Survey on Generative Diffusion Models}.
\newblock \bibinfo{journal}{\emph{IEEE Transactions on Knowledge and Data Engineering}} \bibinfo{volume}{36}, \bibinfo{number}{7} (\bibinfo{year}{2024}), \bibinfo{pages}{2814--2830}.
\newblock
\urldef\tempurl%
\url{https://doi.org/10.1109/TKDE.2024.3361474}
\showDOI{\tempurl}


\bibitem[Carlini et~al\mbox{.}(2023)]%
        {carlini2023extracting}
\bibfield{author}{\bibinfo{person}{Nicholas Carlini}, \bibinfo{person}{Jamie Hayes}, \bibinfo{person}{Milad Nasr}, \bibinfo{person}{Matthew Jagielski}, \bibinfo{person}{Vikash Sehwag}, \bibinfo{person}{Florian Tramèr}, \bibinfo{person}{Borja Balle}, \bibinfo{person}{Daphne Ippolito}, {and} \bibinfo{person}{Eric Wallace}.} \bibinfo{year}{2023}\natexlab{}.
\newblock \bibinfo{title}{Extracting Training Data from Diffusion Models}.
\newblock
\newblock
\showeprint[arxiv]{2301.13188}~[cs.CR]


\bibitem[Carter(1977)]%
        {Sid&Marty_1977}
\bibfield{author}{\bibinfo{person}{Carter}.} \bibinfo{year}{1977}\natexlab{}.
\newblock \bibinfo{booktitle}{\emph{Sid \& Marty Krofft Television Productions Inc. v. McDonald's Corp.}}
\newblock


\bibitem[Dogoulis et~al\mbox{.}(2023)]%
        {dogoulis2023improving}
\bibfield{author}{\bibinfo{person}{Pantelis Dogoulis}, \bibinfo{person}{Giorgos Kordopatis-Zilos}, \bibinfo{person}{Ioannis Kompatsiaris}, {and} \bibinfo{person}{Symeon Papadopoulos}.} \bibinfo{year}{2023}\natexlab{}.
\newblock \showarticletitle{Improving synthetically generated image detection in cross-concept settings}. In \bibinfo{booktitle}{\emph{Proceedings of the 2nd ACM International Workshop on Multimedia AI against Disinformation}}. \bibinfo{pages}{28--35}.
\newblock


\bibitem[D\"{u}tting et~al\mbox{.}(2024)]%
        {Mechanism_design_LLM}
\bibfield{author}{\bibinfo{person}{Paul D\"{u}tting}, \bibinfo{person}{Vahab Mirrokni}, \bibinfo{person}{Renato Paes~Leme}, \bibinfo{person}{Haifeng Xu}, {and} \bibinfo{person}{Song Zuo}.} \bibinfo{year}{2024}\natexlab{}.
\newblock \showarticletitle{Mechanism Design for Large Language Models}. In \bibinfo{booktitle}{\emph{Proceedings of the ACM Web Conference 2024}} (Singapore, Singapore) \emph{(\bibinfo{series}{WWW '24})}. \bibinfo{publisher}{Association for Computing Machinery}, \bibinfo{address}{New York, NY, USA}, \bibinfo{pages}{144–155}.
\newblock
\showISBNx{9798400701719}
\urldef\tempurl%
\url{https://doi.org/10.1145/3589334.3645511}
\showDOI{\tempurl}


\bibitem[Epstein et~al\mbox{.}(2023)]%
        {epstein2023online}
\bibfield{author}{\bibinfo{person}{David~C Epstein}, \bibinfo{person}{Ishan Jain}, \bibinfo{person}{Oliver Wang}, {and} \bibinfo{person}{Richard Zhang}.} \bibinfo{year}{2023}\natexlab{}.
\newblock \showarticletitle{Online detection of ai-generated images}. In \bibinfo{booktitle}{\emph{Proceedings of the IEEE/CVF International Conference on Computer Vision}}. \bibinfo{pages}{382--392}.
\newblock


\bibitem[Gandikota et~al\mbox{.}(2023)]%
        {gandikota2023erasing}
\bibfield{author}{\bibinfo{person}{Rohit Gandikota}, \bibinfo{person}{Joanna Materzynska}, \bibinfo{person}{Jaden Fiotto-Kaufman}, {and} \bibinfo{person}{David Bau}.} \bibinfo{year}{2023}\natexlab{}.
\newblock \bibinfo{title}{Erasing Concepts from Diffusion Models}.
\newblock
\newblock
\showeprint[arxiv]{2303.07345}~[cs.CV]


\bibitem[Gao et~al\mbox{.}(2023)]%
        {gao2023implicit}
\bibfield{author}{\bibinfo{person}{Sicheng Gao}, \bibinfo{person}{Xuhui Liu}, \bibinfo{person}{Bohan Zeng}, \bibinfo{person}{Sheng Xu}, \bibinfo{person}{Yanjing Li}, \bibinfo{person}{Xiaoyan Luo}, \bibinfo{person}{Jianzhuang Liu}, \bibinfo{person}{Xiantong Zhen}, {and} \bibinfo{person}{Baochang Zhang}.} \bibinfo{year}{2023}\natexlab{}.
\newblock \bibinfo{title}{Implicit Diffusion Models for Continuous Super-Resolution}.
\newblock
\newblock
\showeprint[arxiv]{2303.16491}~[cs.CV]


\bibitem[Gong et~al\mbox{.}(2020)]%
        {gong2020survey}
\bibfield{author}{\bibinfo{person}{Maoguo Gong}, \bibinfo{person}{Yu Xie}, \bibinfo{person}{Ke Pan}, \bibinfo{person}{Kaiyuan Feng}, {and} \bibinfo{person}{Alex~Kai Qin}.} \bibinfo{year}{2020}\natexlab{}.
\newblock \showarticletitle{A survey on differentially private machine learning}.
\newblock \bibinfo{journal}{\emph{IEEE computational intelligence magazine}} \bibinfo{volume}{15}, \bibinfo{number}{2} (\bibinfo{year}{2020}), \bibinfo{pages}{49--64}.
\newblock


\bibitem[Heusel et~al\mbox{.}(2018)]%
        {fid2018}
\bibfield{author}{\bibinfo{person}{Martin Heusel}, \bibinfo{person}{Hubert Ramsauer}, \bibinfo{person}{Thomas Unterthiner}, \bibinfo{person}{Bernhard Nessler}, {and} \bibinfo{person}{Sepp Hochreiter}.} \bibinfo{year}{2018}\natexlab{}.
\newblock \bibinfo{title}{GANs Trained by a Two Time-Scale Update Rule Converge to a Local Nash Equilibrium}.
\newblock
\newblock
\showeprint[arxiv]{1706.08500}~[cs.LG]
\urldef\tempurl%
\url{https://arxiv.org/abs/1706.08500}
\showURL{%
\tempurl}


\bibitem[Ho et~al\mbox{.}(2020)]%
        {ho2020denoising}
\bibfield{author}{\bibinfo{person}{Jonathan Ho}, \bibinfo{person}{Ajay Jain}, {and} \bibinfo{person}{Pieter Abbeel}.} \bibinfo{year}{2020}\natexlab{}.
\newblock \bibinfo{title}{Denoising Diffusion Probabilistic Models}.
\newblock
\newblock
\showeprint[arxiv]{2006.11239}~[cs.LG]


\bibitem[Huang et~al\mbox{.}(2021)]%
        {huang2021unlearnable}
\bibfield{author}{\bibinfo{person}{Hanxun Huang}, \bibinfo{person}{Xingjun Ma}, \bibinfo{person}{Sarah~Monazam Erfani}, \bibinfo{person}{James Bailey}, {and} \bibinfo{person}{Yisen Wang}.} \bibinfo{year}{2021}\natexlab{}.
\newblock \bibinfo{title}{Unlearnable Examples: Making Personal Data Unexploitable}.
\newblock
\newblock
\showeprint[arxiv]{2101.04898}~[cs.LG]


\bibitem[Hutsebaut-Buysse et~al\mbox{.}(2022)]%
        {hierarchical_rl}
\bibfield{author}{\bibinfo{person}{Matthias Hutsebaut-Buysse}, \bibinfo{person}{Kevin Mets}, {and} \bibinfo{person}{Steven Latré}.} \bibinfo{year}{2022}\natexlab{}.
\newblock \showarticletitle{Hierarchical Reinforcement Learning: A Survey and Open Research Challenges}.
\newblock \bibinfo{journal}{\emph{Machine Learning and Knowledge Extraction}} \bibinfo{volume}{4}, \bibinfo{number}{1} (\bibinfo{year}{2022}), \bibinfo{pages}{172--221}.
\newblock
\showISSN{2504-4990}
\urldef\tempurl%
\url{https://doi.org/10.3390/make4010009}
\showDOI{\tempurl}


\bibitem[Jiang et~al\mbox{.}(2023)]%
        {ai_art_impact}
\bibfield{author}{\bibinfo{person}{Harry~H. Jiang}, \bibinfo{person}{Lauren Brown}, \bibinfo{person}{Jessica Cheng}, \bibinfo{person}{Mehtab Khan}, \bibinfo{person}{Abhishek Gupta}, \bibinfo{person}{Deja Workman}, \bibinfo{person}{Alex Hanna}, \bibinfo{person}{Johnathan Flowers}, {and} \bibinfo{person}{Timnit Gebru}.} \bibinfo{year}{2023}\natexlab{}.
\newblock \showarticletitle{AI Art and its Impact on Artists}. In \bibinfo{booktitle}{\emph{Proceedings of the 2023 AAAI/ACM Conference on AI, Ethics, and Society}} (Montr\'{e}al, QC, Canada) \emph{(\bibinfo{series}{AIES '23})}. \bibinfo{publisher}{Association for Computing Machinery}, \bibinfo{address}{New York, NY, USA}, \bibinfo{pages}{363–374}.
\newblock
\showISBNx{9798400702310}
\urldef\tempurl%
\url{https://doi.org/10.1145/3600211.3604681}
\showDOI{\tempurl}


\bibitem[Li(2017)]%
        {li2017deep}
\bibfield{author}{\bibinfo{person}{Yuxi Li}.} \bibinfo{year}{2017}\natexlab{}.
\newblock \showarticletitle{Deep reinforcement learning: An overview}.
\newblock \bibinfo{journal}{\emph{arXiv preprint arXiv:1701.07274}} (\bibinfo{year}{2017}).
\newblock


\bibitem[Liao et~al\mbox{.}(2022)]%
        {liao2022artbench}
\bibfield{author}{\bibinfo{person}{Peiyuan Liao}, \bibinfo{person}{Xiuyu Li}, \bibinfo{person}{Xihui Liu}, {and} \bibinfo{person}{Kurt Keutzer}.} \bibinfo{year}{2022}\natexlab{}.
\newblock \bibinfo{title}{The ArtBench Dataset: Benchmarking Generative Models with Artworks}.
\newblock
\newblock
\showeprint[arxiv]{2206.11404}~[cs.CV]
\urldef\tempurl%
\url{https://arxiv.org/abs/2206.11404}
\showURL{%
\tempurl}


\bibitem[Ma et~al\mbox{.}(2024)]%
        {ma2024dataset}
\bibfield{author}{\bibinfo{person}{Rui Ma}, \bibinfo{person}{Qiang Zhou}, \bibinfo{person}{Bangjun Xiao}, \bibinfo{person}{Yizhu Jin}, \bibinfo{person}{Daquan Zhou}, \bibinfo{person}{Xiuyu Li}, \bibinfo{person}{Aishani Singh}, \bibinfo{person}{Yi Qu}, \bibinfo{person}{Kurt Keutzer}, \bibinfo{person}{Xiaodong Xie}, \bibinfo{person}{Jingtong Hu}, \bibinfo{person}{Zhen Dong}, {and} \bibinfo{person}{Shanghang Zhang}.} \bibinfo{year}{2024}\natexlab{}.
\newblock \bibinfo{title}{A Dataset and Benchmark for Copyright Protection from Text-to-Image Diffusion Models}.
\newblock
\newblock
\showeprint[arxiv]{2403.12052}~[cs.CV]


\bibitem[Mnih et~al\mbox{.}(2015)]%
        {DQN2015}
\bibfield{author}{\bibinfo{person}{Volodymyr Mnih}, \bibinfo{person}{Koray Kavukcuoglu}, \bibinfo{person}{David Silver}, \bibinfo{person}{Alex Graves}, \bibinfo{person}{Ioannis Antonoglou}, \bibinfo{person}{Daan Wierstra}, {and} \bibinfo{person}{Martin Riedmiller}.} \bibinfo{year}{2015}\natexlab{}.
\newblock \showarticletitle{Human-level control through deep reinforcement learning}.
\newblock \bibinfo{journal}{\emph{Nature}} \bibinfo{volume}{518}, \bibinfo{number}{7540} (\bibinfo{year}{2015}), \bibinfo{pages}{529--533}.
\newblock
\urldef\tempurl%
\url{https://doi.org/10.1038/nature14236}
\showDOI{\tempurl}


\bibitem[Nguyen et~al\mbox{.}(2022)]%
        {nguyen2022survey}
\bibfield{author}{\bibinfo{person}{Thanh~Tam Nguyen}, \bibinfo{person}{Thanh~Trung Huynh}, \bibinfo{person}{Phi~Le Nguyen}, \bibinfo{person}{Alan Wee-Chung Liew}, \bibinfo{person}{Hongzhi Yin}, {and} \bibinfo{person}{Quoc Viet~Hung Nguyen}.} \bibinfo{year}{2022}\natexlab{}.
\newblock \bibinfo{title}{A Survey of Machine Unlearning}.
\newblock
\newblock
\showeprint[arxiv]{2209.02299}~[cs.LG]


\bibitem[Office(1989)]%
        {Copyrightlaw1989}
\bibfield{author}{\bibinfo{person}{US~Copyright Office}.} \bibinfo{year}{1989}\natexlab{}.
\newblock \bibinfo{title}{Copyright Act 1989}.
\newblock
\newblock


\bibitem[Park et~al\mbox{.}(2023)]%
        {park2023trak}
\bibfield{author}{\bibinfo{person}{Sung~Min Park}, \bibinfo{person}{Kristian Georgiev}, \bibinfo{person}{Andrew Ilyas}, \bibinfo{person}{Guillaume Leclerc}, {and} \bibinfo{person}{Aleksander Madry}.} \bibinfo{year}{2023}\natexlab{}.
\newblock \bibinfo{title}{TRAK: Attributing Model Behavior at Scale}.
\newblock
\newblock
\showeprint[arxiv]{2303.14186}~[stat.ML]


\bibitem[Parliament and Council(2001)]%
        {EUR}
\bibfield{author}{\bibinfo{person}{European Parliament} {and} \bibinfo{person}{Council}.} \bibinfo{year}{2001}\natexlab{}.
\newblock \bibinfo{title}{DIRECTIVE 2001/29/EC OF THE EUROPEAN PARLIAMENT AND OF THE COUNCIL of 22 May 2001 on the harmonisation of certain aspects of copyright and related rights in the information society}.
\newblock
\newblock
\urldef\tempurl%
\url{https://eur-lex.europa.eu/legal-content/EN/ALL/?uri=CELEX%3A32001L0029}
\showURL{%
\tempurl}


\bibitem[Podell et~al\mbox{.}(2023)]%
        {sdxl}
\bibfield{author}{\bibinfo{person}{Dustin Podell}, \bibinfo{person}{Zion English}, \bibinfo{person}{Kyle Lacey}, \bibinfo{person}{Andreas Blattmann}, \bibinfo{person}{Tim Dockhorn}, \bibinfo{person}{Jonas Müller}, \bibinfo{person}{Joe Penna}, {and} \bibinfo{person}{Robin Rombach}.} \bibinfo{year}{2023}\natexlab{}.
\newblock \bibinfo{title}{SDXL: Improving Latent Diffusion Models for High-Resolution Image Synthesis}.
\newblock
\newblock
\showeprint[arxiv]{2307.01952}~[cs.CV]
\urldef\tempurl%
\url{https://arxiv.org/abs/2307.01952}
\showURL{%
\tempurl}


\bibitem[Radford et~al\mbox{.}(2021)]%
        {radford2021learning}
\bibfield{author}{\bibinfo{person}{Alec Radford}, \bibinfo{person}{Jong~Wook Kim}, \bibinfo{person}{Chris Hallacy}, \bibinfo{person}{Aditya Ramesh}, \bibinfo{person}{Gabriel Goh}, \bibinfo{person}{Sandhini Agarwal}, \bibinfo{person}{Girish Sastry}, \bibinfo{person}{Amanda Askell}, \bibinfo{person}{Pamela Mishkin}, \bibinfo{person}{Jack Clark}, \bibinfo{person}{Gretchen Krueger}, {and} \bibinfo{person}{Ilya Sutskever}.} \bibinfo{year}{2021}\natexlab{}.
\newblock \bibinfo{title}{Learning Transferable Visual Models From Natural Language Supervision}.
\newblock
\newblock
\showeprint[arxiv]{2103.00020}~[cs.CV]


\bibitem[Ramesh et~al\mbox{.}(2021)]%
        {dalle}
\bibfield{author}{\bibinfo{person}{Aditya Ramesh}, \bibinfo{person}{Mikhail Pavlov}, \bibinfo{person}{Gabriel Goh}, \bibinfo{person}{Scott Gray}, \bibinfo{person}{Chelsea Voss}, \bibinfo{person}{Alec Radford}, \bibinfo{person}{Mark Chen}, {and} \bibinfo{person}{Ilya Sutskever}.} \bibinfo{year}{2021}\natexlab{}.
\newblock \bibinfo{title}{Zero-Shot Text-to-Image Generation}.
\newblock
\newblock
\showeprint[arxiv]{2102.12092}~[cs.CV]
\urldef\tempurl%
\url{https://arxiv.org/abs/2102.12092}
\showURL{%
\tempurl}


\bibitem[Reed et~al\mbox{.}(2016)]%
        {reed2016generative}
\bibfield{author}{\bibinfo{person}{Scott Reed}, \bibinfo{person}{Zeynep Akata}, \bibinfo{person}{Xinchen Yan}, \bibinfo{person}{Lajanugen Logeswaran}, \bibinfo{person}{Bernt Schiele}, {and} \bibinfo{person}{Honglak Lee}.} \bibinfo{year}{2016}\natexlab{}.
\newblock \showarticletitle{Generative Adversarial Text to Image Synthesis}. In \bibinfo{booktitle}{\emph{Proceedings of the 33rd International Conference on Machine Learning (ICML)}}. \bibinfo{pages}{1060--1069}.
\newblock
\urldef\tempurl%
\url{https://proceedings.mlr.press/v48/reed16.html}
\showURL{%
\tempurl}


\bibitem[Ren et~al\mbox{.}(2024)]%
        {copyright_in_generative_ai}
\bibfield{author}{\bibinfo{person}{Jie Ren}, \bibinfo{person}{Han Xu}, \bibinfo{person}{Pengfei He}, \bibinfo{person}{Yingqian Cui}, \bibinfo{person}{Shenglai Zeng}, \bibinfo{person}{Jiankun Zhang}, \bibinfo{person}{Hongzhi Wen}, \bibinfo{person}{Jiayuan Ding}, \bibinfo{person}{Pei Huang}, \bibinfo{person}{Lingjuan Lyu}, \bibinfo{person}{Hui Liu}, \bibinfo{person}{Yi Chang}, {and} \bibinfo{person}{Jiliang Tang}.} \bibinfo{year}{2024}\natexlab{}.
\newblock \bibinfo{title}{Copyright Protection in Generative AI: A Technical Perspective}.
\newblock
\newblock
\showeprint[arxiv]{2402.02333}~[cs.CR]
\urldef\tempurl%
\url{https://arxiv.org/abs/2402.02333}
\showURL{%
\tempurl}


\bibitem[Rombach et~al\mbox{.}(2022a)]%
        {ldm}
\bibfield{author}{\bibinfo{person}{Robin Rombach}, \bibinfo{person}{Andreas Blattmann}, \bibinfo{person}{Dominik Lorenz}, \bibinfo{person}{Patrick Esser}, {and} \bibinfo{person}{Björn Ommer}.} \bibinfo{year}{2022}\natexlab{a}.
\newblock \bibinfo{title}{High-Resolution Image Synthesis with Latent Diffusion Models}.
\newblock
\newblock
\showeprint[arxiv]{2112.10752}~[cs.CV]
\urldef\tempurl%
\url{https://arxiv.org/abs/2112.10752}
\showURL{%
\tempurl}


\bibitem[Rombach et~al\mbox{.}(2022b)]%
        {diffusion_model}
\bibfield{author}{\bibinfo{person}{Robin Rombach}, \bibinfo{person}{Andreas Blattmann}, \bibinfo{person}{Dominik Lorenz}, \bibinfo{person}{Patrick Esser}, {and} \bibinfo{person}{Björn Ommer}.} \bibinfo{year}{2022}\natexlab{b}.
\newblock \bibinfo{title}{High-Resolution Image Synthesis with Latent Diffusion Models}.
\newblock
\newblock
\showeprint[arxiv]{2112.10752}~[cs.CV]
\urldef\tempurl%
\url{https://arxiv.org/abs/2112.10752}
\showURL{%
\tempurl}


\bibitem[Schulman et~al\mbox{.}(2017)]%
        {schulman2017ppo}
\bibfield{author}{\bibinfo{person}{John Schulman}, \bibinfo{person}{Felix Wolski}, \bibinfo{person}{Prafulla Dhariwal}, \bibinfo{person}{Alec Radford}, {and} \bibinfo{person}{Oleg Klimov}.} \bibinfo{year}{2017}\natexlab{}.
\newblock \showarticletitle{Proximal Policy Optimization Algorithms}.
\newblock \bibinfo{journal}{\emph{arXiv preprint arXiv:1707.06347}} (\bibinfo{year}{2017}).
\newblock
\urldef\tempurl%
\url{https://arxiv.org/abs/1707.06347}
\showURL{%
\tempurl}


\bibitem[Somepalli et~al\mbox{.}(2022)]%
        {somepalli2022diffusion}
\bibfield{author}{\bibinfo{person}{Gowthami Somepalli}, \bibinfo{person}{Vasu Singla}, \bibinfo{person}{Micah Goldblum}, \bibinfo{person}{Jonas Geiping}, {and} \bibinfo{person}{Tom Goldstein}.} \bibinfo{year}{2022}\natexlab{}.
\newblock \bibinfo{title}{Diffusion Art or Digital Forgery? Investigating Data Replication in Diffusion Models}.
\newblock
\newblock
\showeprint[arxiv]{2212.03860}~[cs.LG]


\bibitem[Vombatkere et~al\mbox{.}(2024)]%
        {vombatkere2024tiktokart}
\bibfield{author}{\bibinfo{person}{Karan Vombatkere}, \bibinfo{person}{Sepehr Mousavi}, \bibinfo{person}{Savvas Zannettou}, \bibinfo{person}{Franziska Roesner}, {and} \bibinfo{person}{Krishna~P. Gummadi}.} \bibinfo{year}{2024}\natexlab{}.
\newblock \bibinfo{title}{TikTok and the Art of Personalization: Investigating Exploration and Exploitation on Social Media Feeds}.
\newblock
\newblock
\showeprint[arxiv]{2403.12410}~[cs.SI]
\urldef\tempurl%
\url{https://arxiv.org/abs/2403.12410}
\showURL{%
\tempurl}


\bibitem[Vyas et~al\mbox{.}(2023)]%
        {vyas2023provable}
\bibfield{author}{\bibinfo{person}{Nikhil Vyas}, \bibinfo{person}{Sham Kakade}, {and} \bibinfo{person}{Boaz Barak}.} \bibinfo{year}{2023}\natexlab{}.
\newblock \bibinfo{title}{On Provable Copyright Protection for Generative Models}.
\newblock
\newblock
\showeprint[arxiv]{2302.10870}~[cs.LG]


\bibitem[Wang et~al\mbox{.}(2024a)]%
        {economicSolutionCopyright}
\bibfield{author}{\bibinfo{person}{Jiachen~T. Wang}, \bibinfo{person}{Zhun Deng}, \bibinfo{person}{Hiroaki Chiba-Okabe}, \bibinfo{person}{Boaz Barak}, {and} \bibinfo{person}{Weijie~J. Su}.} \bibinfo{year}{2024}\natexlab{a}.
\newblock \bibinfo{title}{An Economic Solution to Copyright Challenges of Generative AI}.
\newblock
\newblock
\showeprint[arxiv]{2404.13964}~[cs.LG]
\urldef\tempurl%
\url{https://arxiv.org/abs/2404.13964}
\showURL{%
\tempurl}


\bibitem[Wang et~al\mbox{.}(2024b)]%
        {wang2024comprehensivesurveycontinuallearning}
\bibfield{author}{\bibinfo{person}{Liyuan Wang}, \bibinfo{person}{Xingxing Zhang}, \bibinfo{person}{Hang Su}, {and} \bibinfo{person}{Jun Zhu}.} \bibinfo{year}{2024}\natexlab{b}.
\newblock \bibinfo{title}{A Comprehensive Survey of Continual Learning: Theory, Method and Application}.
\newblock
\newblock
\showeprint[arxiv]{2302.00487}~[cs.LG]
\urldef\tempurl%
\url{https://arxiv.org/abs/2302.00487}
\showURL{%
\tempurl}


\bibitem[Wu et~al\mbox{.}(2023)]%
        {aigc}
\bibfield{author}{\bibinfo{person}{Jiayang Wu}, \bibinfo{person}{Wensheng Gan}, \bibinfo{person}{Zefeng Chen}, \bibinfo{person}{Shicheng Wan}, {and} \bibinfo{person}{Hong Lin}.} \bibinfo{year}{2023}\natexlab{}.
\newblock \bibinfo{title}{AI-Generated Content (AIGC): A Survey}.
\newblock
\newblock
\showeprint[arxiv]{2304.06632}~[cs.AI]
\urldef\tempurl%
\url{https://arxiv.org/abs/2304.06632}
\showURL{%
\tempurl}


\bibitem[Wu et~al\mbox{.}(2024)]%
        {wu2024smalllanguagemodelsserve}
\bibfield{author}{\bibinfo{person}{Xuansheng Wu}, \bibinfo{person}{Huachi Zhou}, \bibinfo{person}{Yucheng Shi}, \bibinfo{person}{Wenlin Yao}, \bibinfo{person}{Xiao Huang}, {and} \bibinfo{person}{Ninghao Liu}.} \bibinfo{year}{2024}\natexlab{}.
\newblock \bibinfo{title}{Could Small Language Models Serve as Recommenders? Towards Data-centric Cold-start Recommendations}.
\newblock
\newblock
\showeprint[arxiv]{2306.17256}~[cs.IR]
\urldef\tempurl%
\url{https://arxiv.org/abs/2306.17256}
\showURL{%
\tempurl}


\bibitem[Xu et~al\mbox{.}(2018)]%
        {xu2018attngan}
\bibfield{author}{\bibinfo{person}{Tao Xu}, \bibinfo{person}{Pengchuan Zhang}, \bibinfo{person}{Qiuyuan Huang}, \bibinfo{person}{Han Zhang}, \bibinfo{person}{Zhe Gan}, \bibinfo{person}{Xiaolei Huang}, {and} \bibinfo{person}{Xiaodong He}.} \bibinfo{year}{2018}\natexlab{}.
\newblock \showarticletitle{AttnGAN: Fine-Grained Text to Image Generation with Attentional Generative Adversarial Networks}. In \bibinfo{booktitle}{\emph{Proceedings of the IEEE Conference on Computer Vision and Pattern Recognition (CVPR)}}. \bibinfo{pages}{1316--1324}.
\newblock
\urldef\tempurl%
\url{https://doi.org/10.1109/CVPR.2018.00143}
\showDOI{\tempurl}


\bibitem[Yang et~al\mbox{.}(2024)]%
        {diffusion_model_survey}
\bibfield{author}{\bibinfo{person}{Ling Yang}, \bibinfo{person}{Zhilong Zhang}, \bibinfo{person}{Yang Song}, \bibinfo{person}{Shenda Hong}, \bibinfo{person}{Runsheng Xu}, \bibinfo{person}{Yue Zhao}, \bibinfo{person}{Wentao Zhang}, \bibinfo{person}{Bin Cui}, {and} \bibinfo{person}{Ming-Hsuan Yang}.} \bibinfo{year}{2024}\natexlab{}.
\newblock \bibinfo{title}{Diffusion Models: A Comprehensive Survey of Methods and Applications}.
\newblock
\newblock
\showeprint[arxiv]{2209.00796}~[cs.LG]
\urldef\tempurl%
\url{https://arxiv.org/abs/2209.00796}
\showURL{%
\tempurl}


\bibitem[Zhang et~al\mbox{.}(2023a)]%
        {zhang2023forgetmenot}
\bibfield{author}{\bibinfo{person}{Eric Zhang}, \bibinfo{person}{Kai Wang}, \bibinfo{person}{Xingqian Xu}, \bibinfo{person}{Zhangyang Wang}, {and} \bibinfo{person}{Humphrey Shi}.} \bibinfo{year}{2023}\natexlab{a}.
\newblock \bibinfo{title}{Forget-Me-Not: Learning to Forget in Text-to-Image Diffusion Models}.
\newblock
\newblock
\showeprint[arxiv]{2303.17591}~[cs.CV]


\bibitem[Zhang et~al\mbox{.}(2023b)]%
        {zhang2023forgotmenot}
\bibfield{author}{\bibinfo{person}{Eric Zhang}, \bibinfo{person}{Kai Wang}, \bibinfo{person}{Xingqian Xu}, \bibinfo{person}{Zhangyang Wang}, {and} \bibinfo{person}{Humphrey Shi}.} \bibinfo{year}{2023}\natexlab{b}.
\newblock \bibinfo{title}{Forget-Me-Not: Learning to Forget in Text-to-Image Diffusion Models}.
\newblock
\newblock
\showeprint[arxiv]{2303.17591}~[cs.CV]
\urldef\tempurl%
\url{https://arxiv.org/abs/2303.17591}
\showURL{%
\tempurl}


\bibitem[Zhang et~al\mbox{.}(2017)]%
        {stackgan2017}
\bibfield{author}{\bibinfo{person}{Han Zhang}, \bibinfo{person}{Tao Xu}, \bibinfo{person}{Hongsheng Li}, \bibinfo{person}{Shaoting Zhang}, \bibinfo{person}{Xiaogang Wang}, \bibinfo{person}{Xiaolei Huang}, {and} \bibinfo{person}{Dimitris Metaxas}.} \bibinfo{year}{2017}\natexlab{}.
\newblock \bibinfo{title}{StackGAN: Text to Photo-realistic Image Synthesis with Stacked Generative Adversarial Networks}.
\newblock
\newblock
\showeprint[arxiv]{1612.03242}~[cs.CV]
\urldef\tempurl%
\url{https://arxiv.org/abs/1612.03242}
\showURL{%
\tempurl}


\bibitem[Zhang et~al\mbox{.}(2018)]%
        {zhang2018unreasonable}
\bibfield{author}{\bibinfo{person}{Richard Zhang}, \bibinfo{person}{Phillip Isola}, \bibinfo{person}{Alexei~A. Efros}, \bibinfo{person}{Eli Shechtman}, {and} \bibinfo{person}{Oliver Wang}.} \bibinfo{year}{2018}\natexlab{}.
\newblock \bibinfo{title}{The Unreasonable Effectiveness of Deep Features as a Perceptual Metric}.
\newblock
\newblock
\showeprint[arxiv]{1801.03924}~[cs.CV]


\bibitem[Zheng et~al\mbox{.}(2024)]%
        {zheng2024dtrak}
\bibfield{author}{\bibinfo{person}{Xiaosen Zheng}, \bibinfo{person}{Tianyu Pang}, \bibinfo{person}{Chao Du}, \bibinfo{person}{Jing Jiang}, {and} \bibinfo{person}{Min Lin}.} \bibinfo{year}{2024}\natexlab{}.
\newblock \bibinfo{title}{Intriguing Properties of Data Attribution on Diffusion Models}.
\newblock
\newblock
\showeprint[arxiv]{2311.00500}~[cs.LG]
\urldef\tempurl%
\url{https://arxiv.org/abs/2311.00500}
\showURL{%
\tempurl}


\end{thebibliography}

\end{document}